\documentclass[preprint,12pt]{elsarticle}

\usepackage{amssymb}

\usepackage[utf8]{inputenc} 
\usepackage[T1]{fontenc}    
\usepackage{hyperref}       
\usepackage{url}            
\usepackage{booktabs}       
\usepackage{amsfonts}       
\usepackage{nicefrac}       
\usepackage{microtype}      
\usepackage{verbatim}
\usepackage{enumitem}
\usepackage{longtable}
\usepackage{multirow}
\usepackage[normalem]{ulem}
\usepackage{graphicx, lipsum, comment, fullpage, multicol, color,threeparttable,subcaption}
\usepackage{pdflscape}

\journal{Pattern Recognition}

\begin{document}

\begin{frontmatter}



\title{To Cluster, or Not to Cluster: \\ 
An Analysis of Clusterability Methods}



\author[scu]{Andreas Adolfsson}
\ead{aadolfsson@scu.edu}

\author[scu]{Margareta Ackerman\fnref{fn1}}
\ead{mackerman@scu.edu}

\author[fsu]{Naomi C. Brownstein\fnref{fn1}\corref{cor1}}
\ead{naomi.brownstein@med.fsu.edu}


\cortext[cor1]{Corresponding author}
\fntext[fn1]{These authors contributed equally.}
\address[scu]{Department of Computer Engineering\\ Santa Clara University\\
500 El Camino Real\\
Santa Clara, CA 95053
}
\address[fsu]{Department of Behavioral Sciences and Social Medicine\\
       Florida State University\\ 1115 West Call Street\\      
Tallahassee, FL, 32306-4300}

\begin{abstract}
Clustering is an essential data mining tool that aims to discover inherent cluster structure in data. For most applications, applying clustering is only appropriate when cluster structure is present. As such, the study of clusterability, which evaluates whether data possesses such structure, is an integral part of cluster analysis. However, methods for evaluating clusterability  vary radically, making it challenging to select a suitable measure. In this paper, we perform an extensive comparison of measures of clusterability and provide guidelines that clustering users can reference to select suitable measures for their applications. 
\end{abstract}


\end{frontmatter}


\section{Introduction}\label{intro}

Clustering is an ubiquitous data analysis tool 
applied across diverse disciplines, such as bioinformatics, marketing, and image segmentation. Its wide utility is perhaps unsurprising, as its intuitive aim - to divide data into groups of similar items - applies at various stages of the data analysis process, from exploratory data analysis to collaborative filtering. 

Despite its popularity, we have barely scratched the surface on 
fundamental questions about clustering. Issues as basic as the definition of clustering 
are being raised~\citep{Kleinberg, NIPS2008}. Differences between clustering algorithms are studied 
to decide which should be used under different circumstances~\citep{NIPS2010,COLT2010,IJCAI2011,weighted}. Yet, perhaps an even 
more fundamental issue than algorithm selection 
is 
when clustering should, or should not, be applied. 

Cluster analysis may be utilized with either realistic or constructive aims~\cite{hennig2015true}. The goal of realistic clustering is to uncover real groupings inherent in the data. For instance, phylogenetic analysis and other life science clustering applications look for real groupings in the data, and as such have realistic aims. By contrast, constructive clustering is relevant when clustering should take place irrespective of whether inherent cluster structure is present. Yet, even when the goal of clustering is constructive, for instance, while the application of cluster analysis to market segmentation may be primarily constructive, users may be interested in realistic groups when such are present in the data.


%
%
%
%
%

Clustering with realistic aims, which is our focus here, is only appropriate when cluster structure is present in the data. Otherwise, the results of any clustering technique become necessarily arbitrary and consequently potentially misleading. 
For concreteness, consider a data set 
generated from a single Gaussian distribution. Because the data contains only one cluster, 
further subdivision 
would be artificial. 
Most clustering algorithms (e.g. $k$-means with $k\geq 2$) would find multiple clusters in this data, even though no multi-cluster structure is present. As such, the application of these data mining tools rely on the presence of inherent structure, rendering \emph{notions of clusterability}, which aim to quantify the degree of cluster structure, integral to cluster analysis.  Clusterability analysis should precede the application of clustering algorithms, as the success of any clustering method depends on the presence of underlying cluster structure.

\begin{figure*}
\includegraphics[width=\textwidth,scale=0.35]{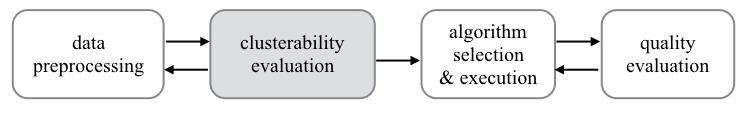}
\caption{Clustering pipeline. This figure shows the feedback pathways in cluster analysis and the role of clusterability in this process.  
}
\label{pipeline}
\end{figure*}

To see how clusterability fits within the clustering process, consider the
 pipeline depicted in Figure~\ref{pipeline}.\footnote{A similar pipeline is presented in the famous survey of clustering algorithms by Xu and Wunsch~\cite{xu2005survey}, sans the second 
step. Figure~\ref{pipeline} shows how clusterability fits within the clustering process.} The process begins with data preprocessing, often involving feature selection or extraction. Next, clusterability analysis determines whether the data possesses inherent cluster structure. If the data does not possess sufficient cluster structure to be meaningfully partitioned, then clustering may not be suitable for the given data, or the data may need to be reprocessed. Otherwise, if the data is found to be clusterable, a 
clustering algorithm may be selected or developed.\footnote{Note that multiple methods should be considered at this step, because different algorithms are apt at identifying different types of cluster structures ~\cite{NIPS2010, ackerman2014incremental}.} Finally, 
the solution is validated by applying clustering quality measures~\cite{NIPS2008, Milligan}, which may result in the selection of an alternate algorithm if a sufficiently high quality clustering has not been found. 


Not unlike clustering algorithms,  
notions of clusterability, summarized in Section \ref{previous}, disagree with each other 
in surprising ways \cite{NIPS2010}. An 
analysis by Ackerman and Ben-David 
\cite{ackerman2009clusterability} reveals that many notions of clusterability are pairwise distinct  - despite the fact that they all attempt to evaluate the same characteristic.
%
The plethora of clusterability methods 
presents a dilemma: 
how should one select a clusterability measure suited to their data?\footnote{The original ``user's dilemma'' refers to the problem of selecting a clustering algorithm for a given task. Selecting a notion of clusterability is 
another dilemma that the clustering user 
faces, addressed in the current work.} Ben-David~
\cite{ben2015computational} approaches the problem from a theoretical standpoint, offering several properties that notions of clusterability should satisfy. In particular, he argues that clusterability notions need to be computationally efficient and effective; see section~\ref{requirements} for details. 


\begin{figure}
\begin{center}
\includegraphics[scale=0.3]{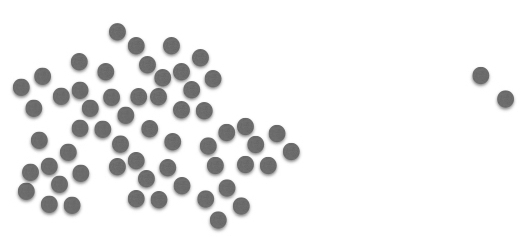}
\end{center}
\caption{Data with ambiguous cluster structure. Outliers can be ignored or considered a small cluster. 
Whether or not this data is considered clusterable depends on the needs of the given application. }\label{outlier_fig}
\end{figure}

The effectiveness requirement is complicated by the inherent ambiguity of cluster analysis. Whether an algorithm is effective for a given application depends on the needs of that applications~\citep{NIPS2010}. Similarly, the needs of the application at hand may dictate whether the given data is clusterable. For example, allowing small clusters can change how we evaluate the clusterability of the data in Figure \ref{outlier_fig}; If small clusters are appropriate for the given application, then the data would be clusterable, whereas otherwise it would be unclusterable. Distant elements are typically considered important when clustering phylogenetic data, whereas outliers are often ignored when clustering is applied to market segmentation; 
a detailed discussion is provided elsewhere \cite{weighted}. Such considerations make room for multiple legitimate clusterability measures, and create the need for guidelines that would help a user determine which notion to choose for their data.

In this paper, we perform an extensive empirical statistical analysis of clusterability measures. 
We not only 
identify effective notions, but also 
discover important differences amongst them that can enable a clustering user to make informed decisions when selecting a clusterability technique.


We begin by formalizing clusterability and presenting several properties that notions of clusterability should satisfy. This enables us to identify promising clusterability measures, which we overview in 
Section \ref{previous} 
We then present our 
extensive simulations, which allow us to discern between approaches to clusterability and determine which are more appropriate under different circumstances. Next, we apply these measures of clusterability to real data. We conclude with a summary of our findings and recommendations. 

\section{Measures of Clusterability}\label{previous}


A plethora for measures of clusterability have been proposed in the literature.  We begin by formalizing clusterability and propose several requirements. This formal framework enables us to focus our analysis on the most promising measures. An extensive statistical analysis of these measures is performed in Section~\ref{sims}. 

\subsection{Requirements of Clusterability Measures}\label{requirements}


While the aim of clustering - to group similar items - is both intuitive and highly applicable, formalizing clustering has proven to be a difficult task. Despite extensive research in the field, clustering remains ill-defined~\citep{NIPS2008}. 
In particular, we do not have a formal definition of clusterability (or even a formal definition of clustering
\footnote{While  axioms and properties have been proposed, e.g. \cite{Kleinberg,NIPS2008,NIPS2010}, we do not yet have a formal definition of clustering, clustering functions, or clusterability.}). Recently, 
Ben-David~
\cite{ben2015computational} began tackling the challenge of formalizing clusterability by proposing several interesting properties. In this section, we 
distill several properties that will help sift through the plethora of clusterability measures 
to identify those that are most likely 
to be useful in practice. Two of our properties, the first and third, are based on Ben-David's requirements. 

A \emph{measure of clusterability} is a function that takes in a data set and outputs a number quantifying its degree of inherent cluster structure.\footnote{
Outputs may be real values, binary indicators (``clusterable'' or ''unclusterable''), or probability measures.
} Naturally, additional requirements are necessary, as functions (e.g. constant functions that declare all data sets to be clusterable) can easily contradict our intuition about how a measure of clusterability should behave. 
To this end, we propose several properties. We rely on these properties to select clusterability measures for our analysis in Section \ref{sims}.



\begin{itemize}
\item \textbf{Efficiency}: \emph{For practical utility, a measure of clusterability should be efficiently 
computed.} Without being overly restrictive, we require that the measures be computable in low polynomial time. In particular, this eliminates measures that are NP-hard to compute. This requirement relates to Ben-David's
third requirement \cite{ben2015computational}.
\item \textbf{Algorithm Independence}: \emph{The clusterability measure should not be based on a specific clustering algorithm or objective function.} Notions of clusterability that are based on a specific algorithm ask a different question than ours; While we ask whether data is clusterable, algorithm-specific notions aim to discover if the data can be clustered using a particular method. 
\item \textbf{Effectiveness}: 
\emph{The measure of clusterability should be highly accurate in identifying data as clusterable or unclusterable.}  As discussed in 
section \ref{intro}, 
the inherent ambiguity of clustering necessitates flexibility on what it means to be ``clusterable.'' Yet, there are many clear examples of both clusterable and unclusterable data, such as, for instance, a single Guassian (unclusterable) or two well-separated Guassians (clusterable). This property is related to the first requirement in Ben-David's paper~
\cite{ben2015computational}.
\end{itemize}

The first 
requirement is that measures be efficient in practice. We address this requirement in section \ref{efficiency}. They should certainly be computable in polynomial time
and run in reasonable time on large data sets. Our second requirement concerns the role of clusterability 
in the clustering pipeline in Figure \ref{pipeline}. Since different clustering algorithms are adept at identifying distinct types of cluster structure~\citep{oligarchies,ackerman2014incremental}, centering a measure of clusterability on a specific algorithm restricts it from identifying structure that the underlying algorithm cannot capture. 


%
%
%
Finally, the third and most challenging requirement, the identification of methods that satisfy effectiveness, 
is the focus of our work. 
We first collect a body of existing measures and propose additional measures that satisfy the first two requirements. Next, we 
empirically compare the performance of these methods on a large number of 
real and simulated data sets. Our data includes many examples that leave no room for ambiguity, allowing us to determine which clusterability measures are effective. Differences in their behavior on more ambiguous data allow us to identify guidelines that can be used to help clustering practitioners select suitable notions of clusterability for their tasks. The combination of extensive simulations and analysis on real data allows us to gain insight into different approaches to clusterability evaluation, compare their effectiveness and fit for distinct clustering scenarios.


\subsection{Effective Approaches to Clusterability Evaluation}

\begin{figure*}
\begin{center}
  \begin{subfigure}[b]{0.3\textwidth}
    \includegraphics[width=\textwidth]{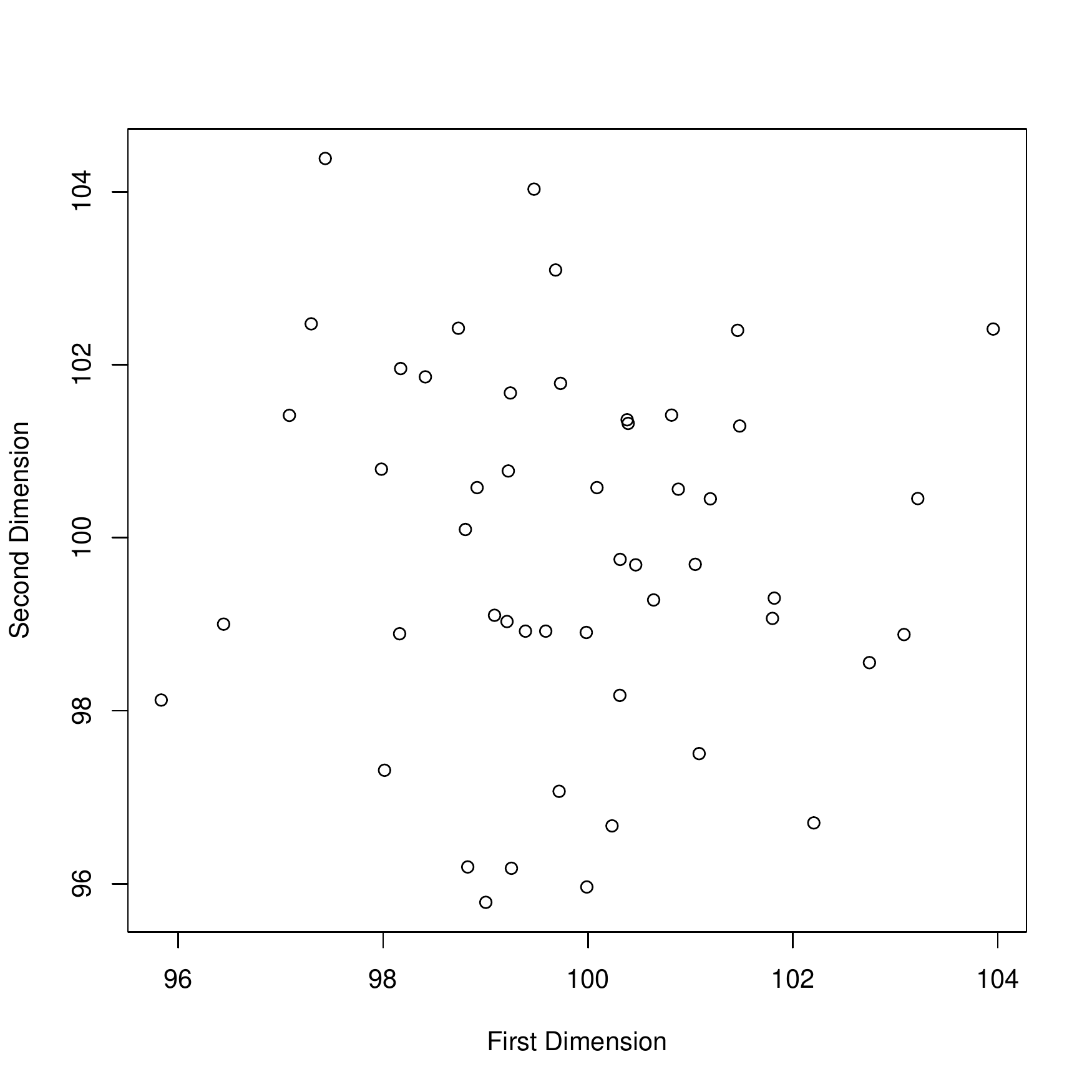}
    \caption{One Cluster: Original Data}
    \label{oneclustoriginal}
  \end{subfigure}
  \begin{subfigure}[b]{0.3\textwidth}
    \includegraphics[width=\textwidth]{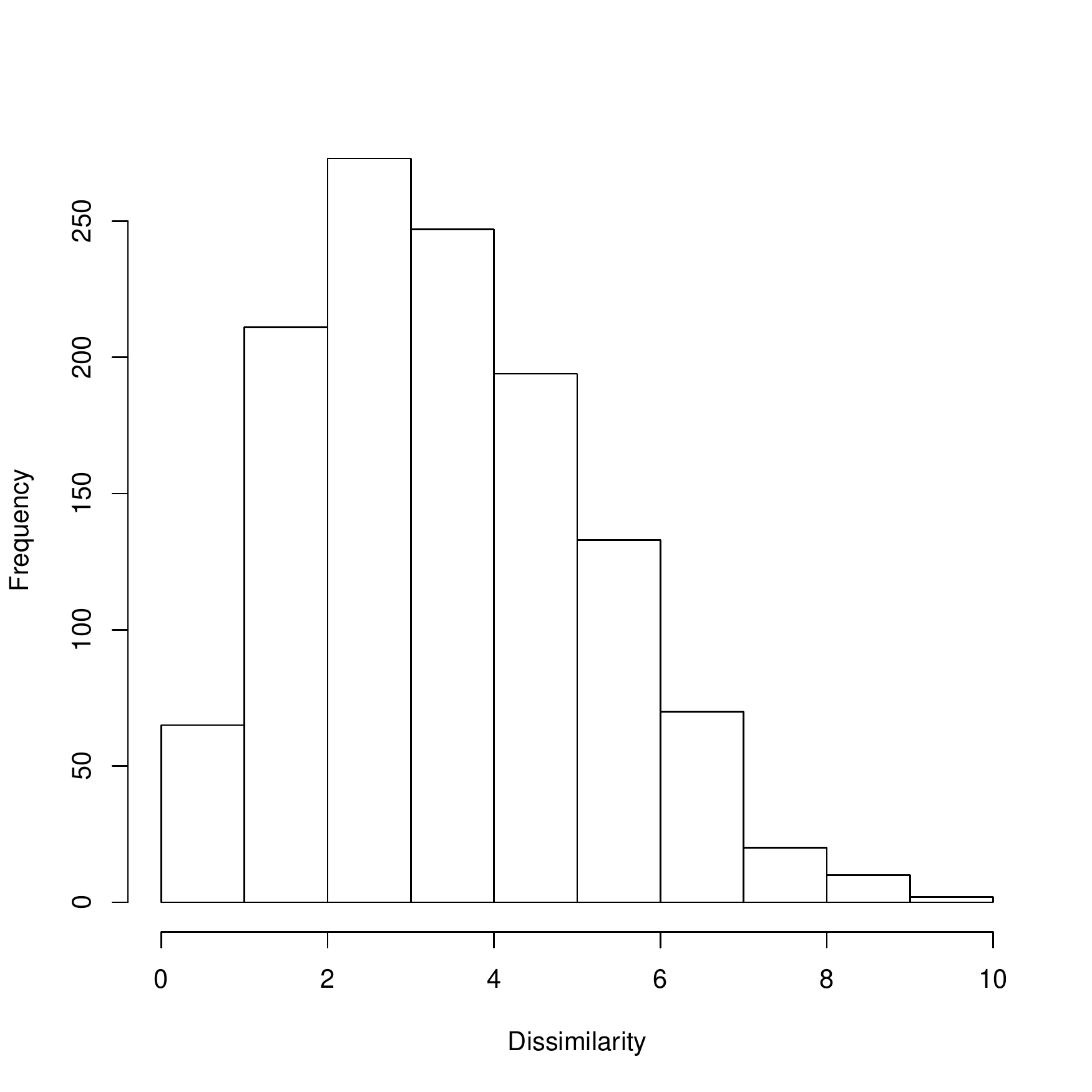}
    \caption{One Cluster: Distances}
    \label{oneclustdist}
  \end{subfigure}
  \begin{subfigure}[b]{0.3\textwidth}
    \includegraphics[width=\textwidth]{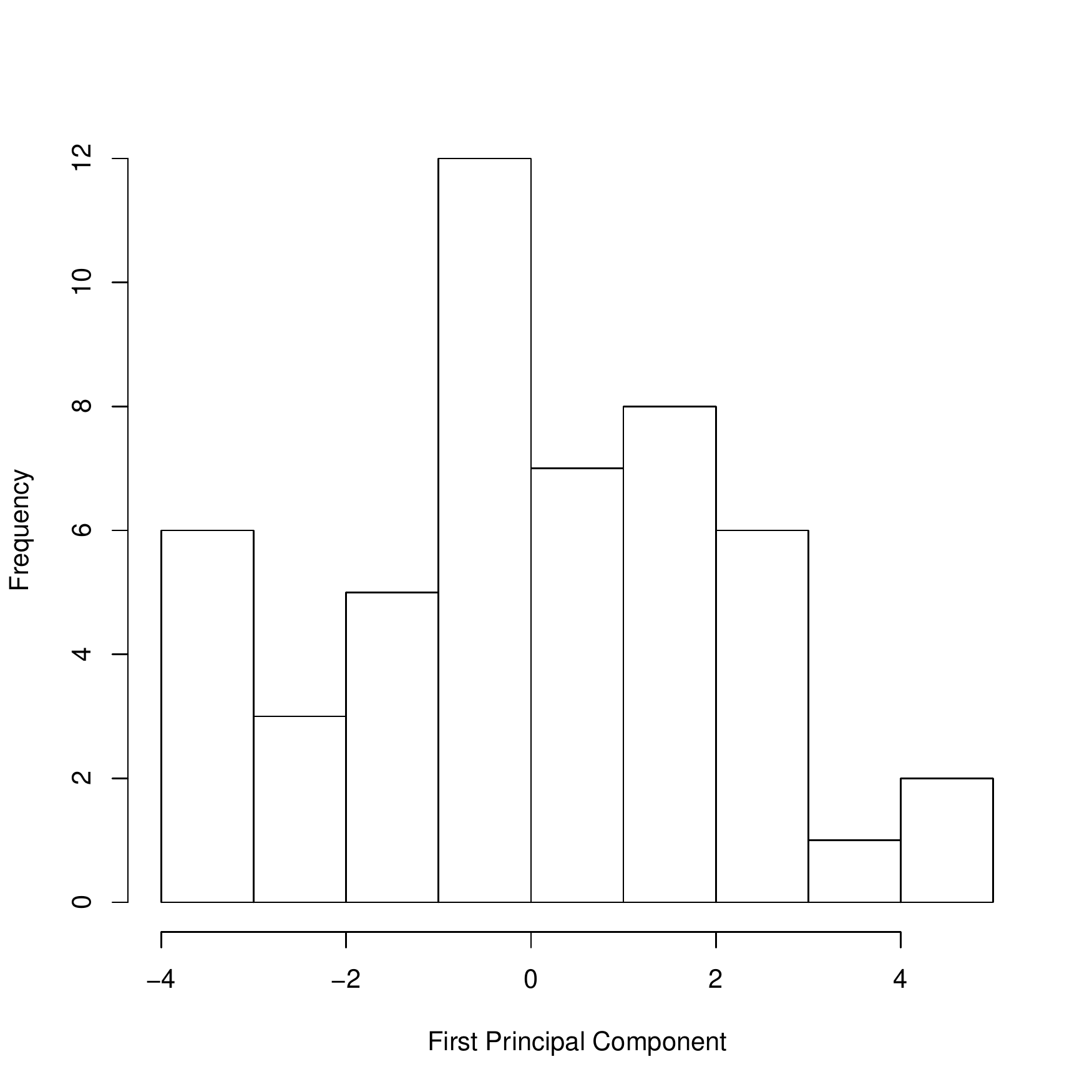}
    \caption{One Cluster: PCA}
    \label{oneclustpca}
  \end{subfigure}
  \\
   \begin{subfigure}[b]{0.3\textwidth}
    \includegraphics[width=\textwidth]{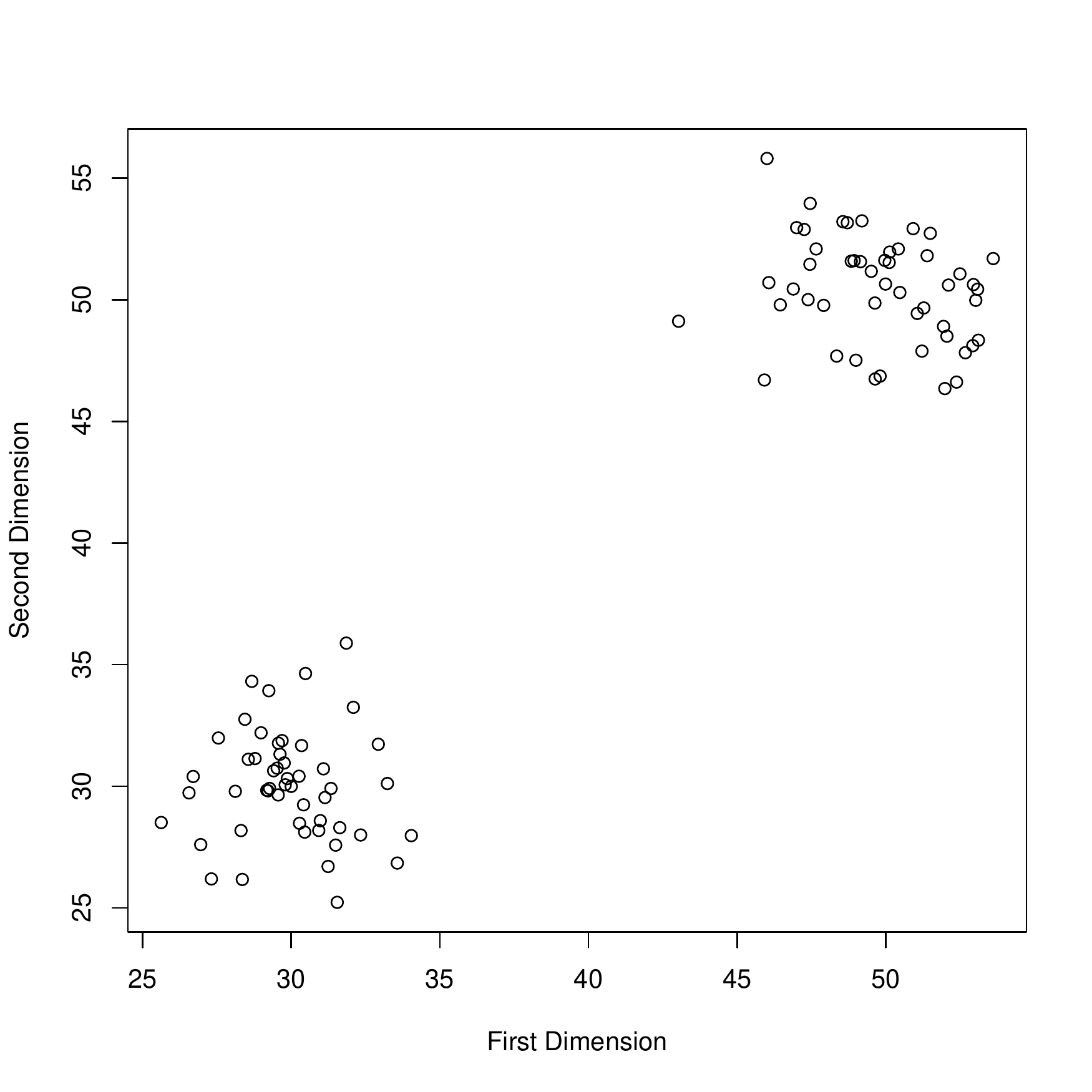}
    \caption{Two Clusters: Original Data}
    \label{twoclustoriginal}
  \end{subfigure}
 \begin{subfigure}[b]{0.3\textwidth}
    \includegraphics[width=\textwidth]{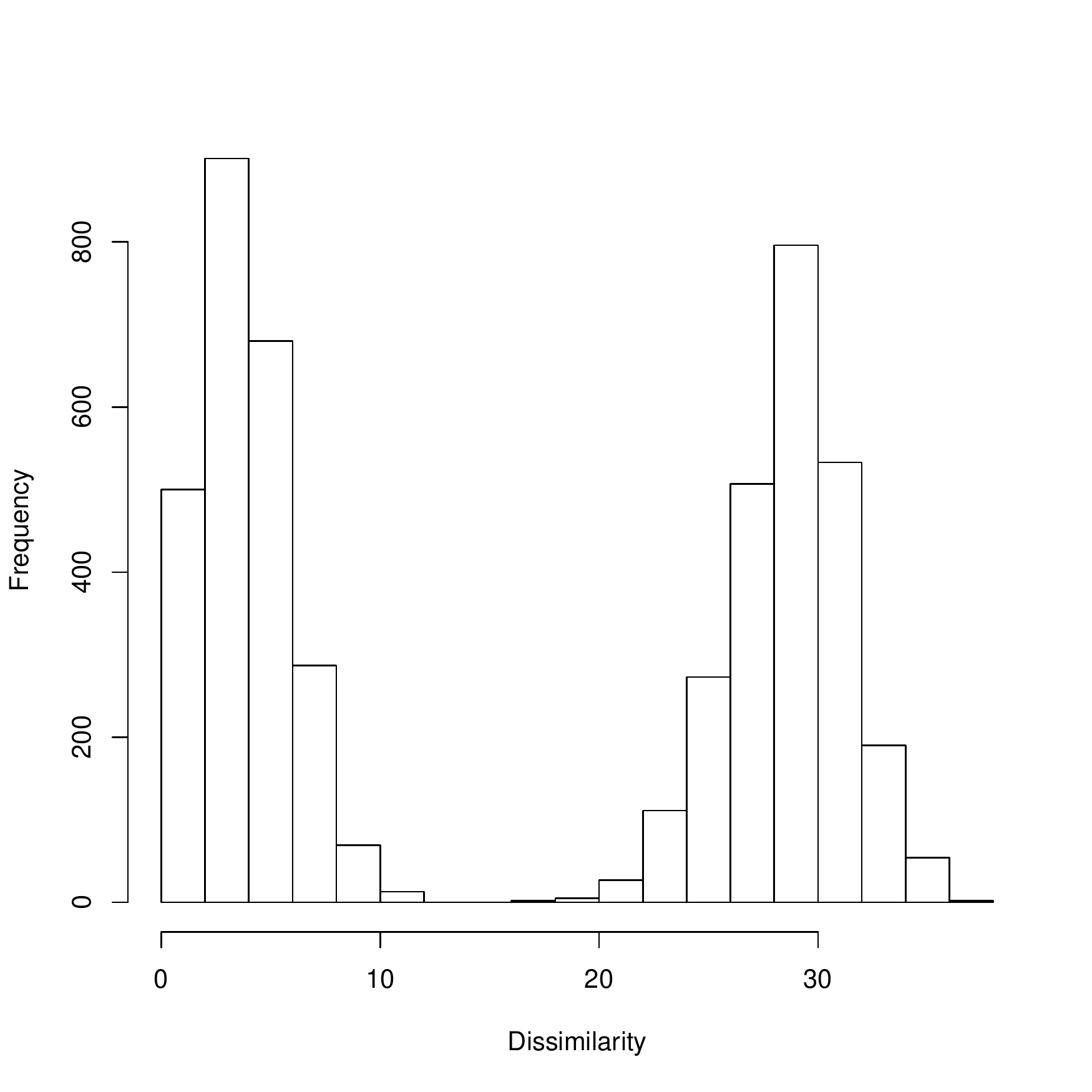}
    \caption{Two Clusters: Distances}
    \label{twoclustdist}
  \end{subfigure}
  \begin{subfigure}[b]{0.3\textwidth}
    \includegraphics[width=\textwidth]{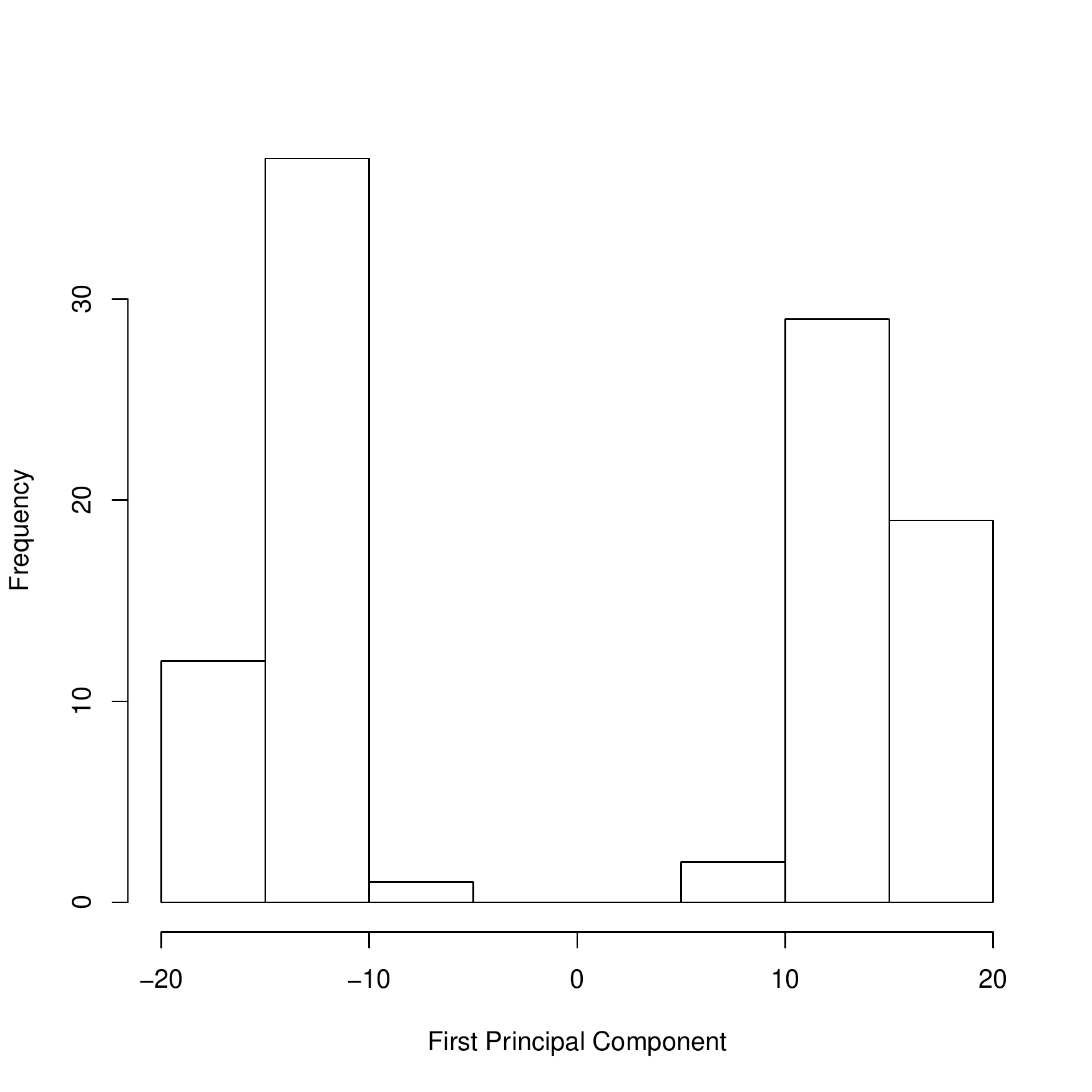}
    \caption{Two Clusters: PCA}
    \label{twoclustpca}
  \end{subfigure}
  \\
  \begin{subfigure}[b]{0.3\textwidth}
    \includegraphics[width=\textwidth]{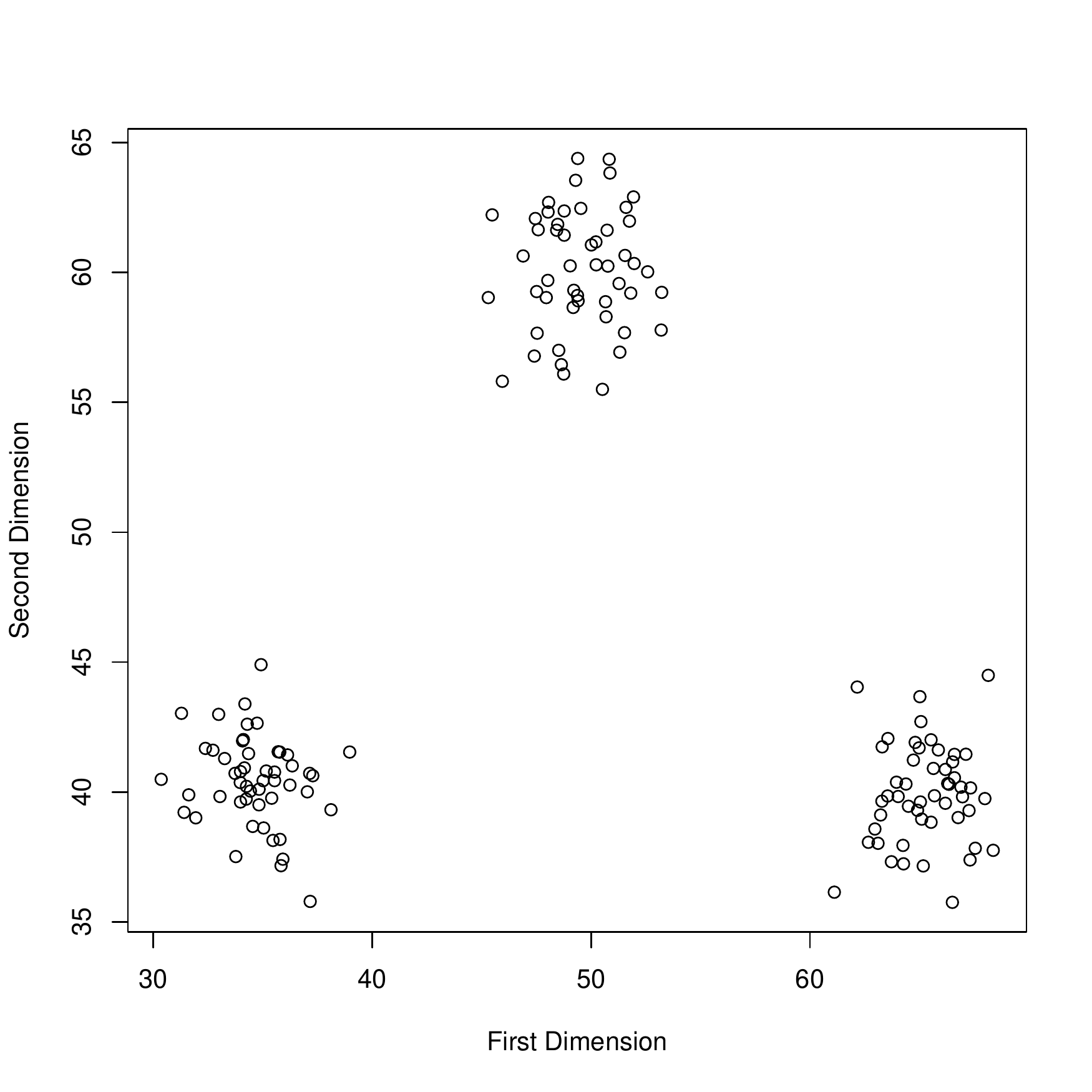}
     \caption{Three Clusters: Original Data}
    \label{threeclustoriginal}
  \end{subfigure}
 \begin{subfigure}[b]{0.3\textwidth}
    \includegraphics[width=\textwidth]{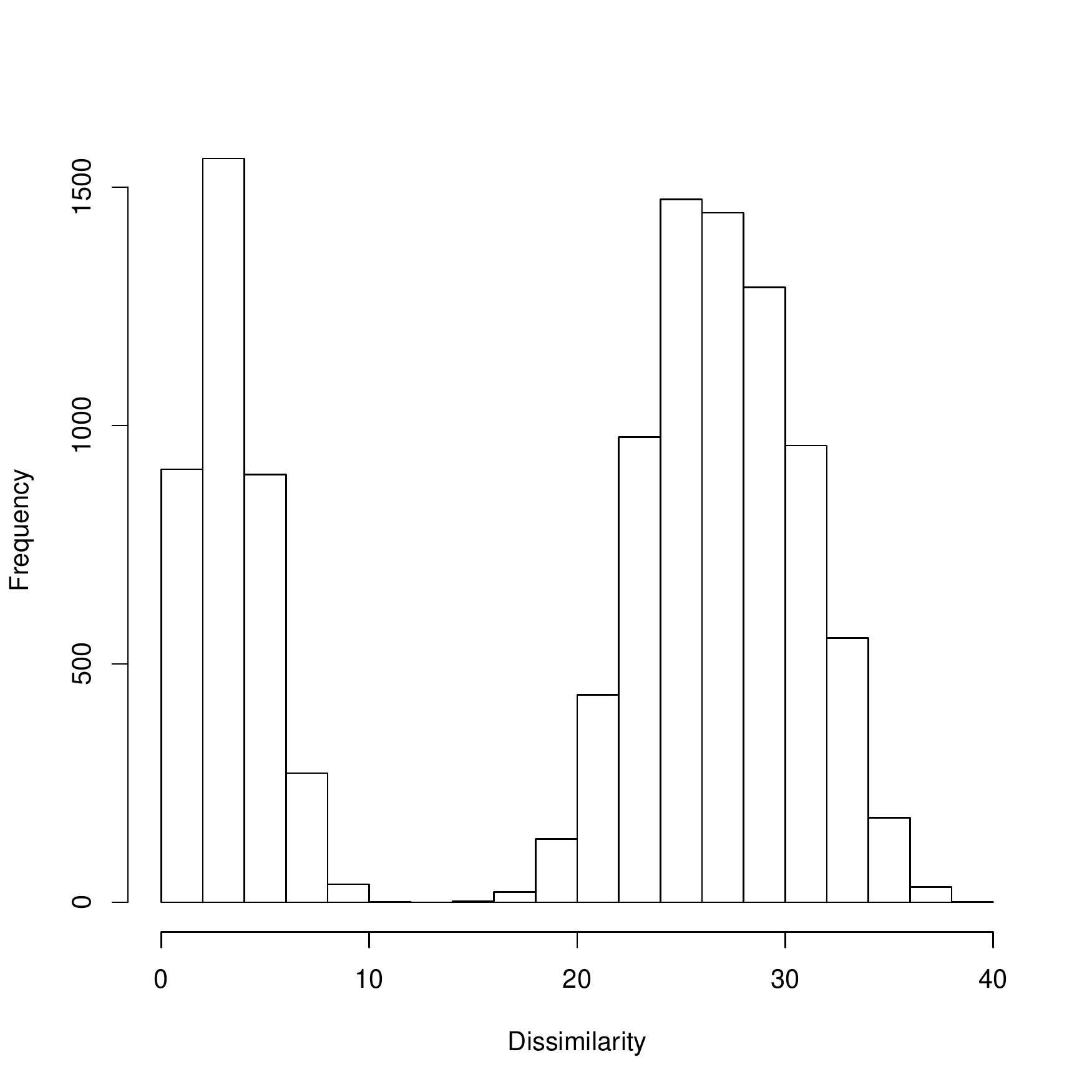}
    \caption{Three Clusters: Distances}
    \label{threeclustdist}
  \end{subfigure}
  \begin{subfigure}[b]{0.3\textwidth}
    \includegraphics[width=\textwidth]{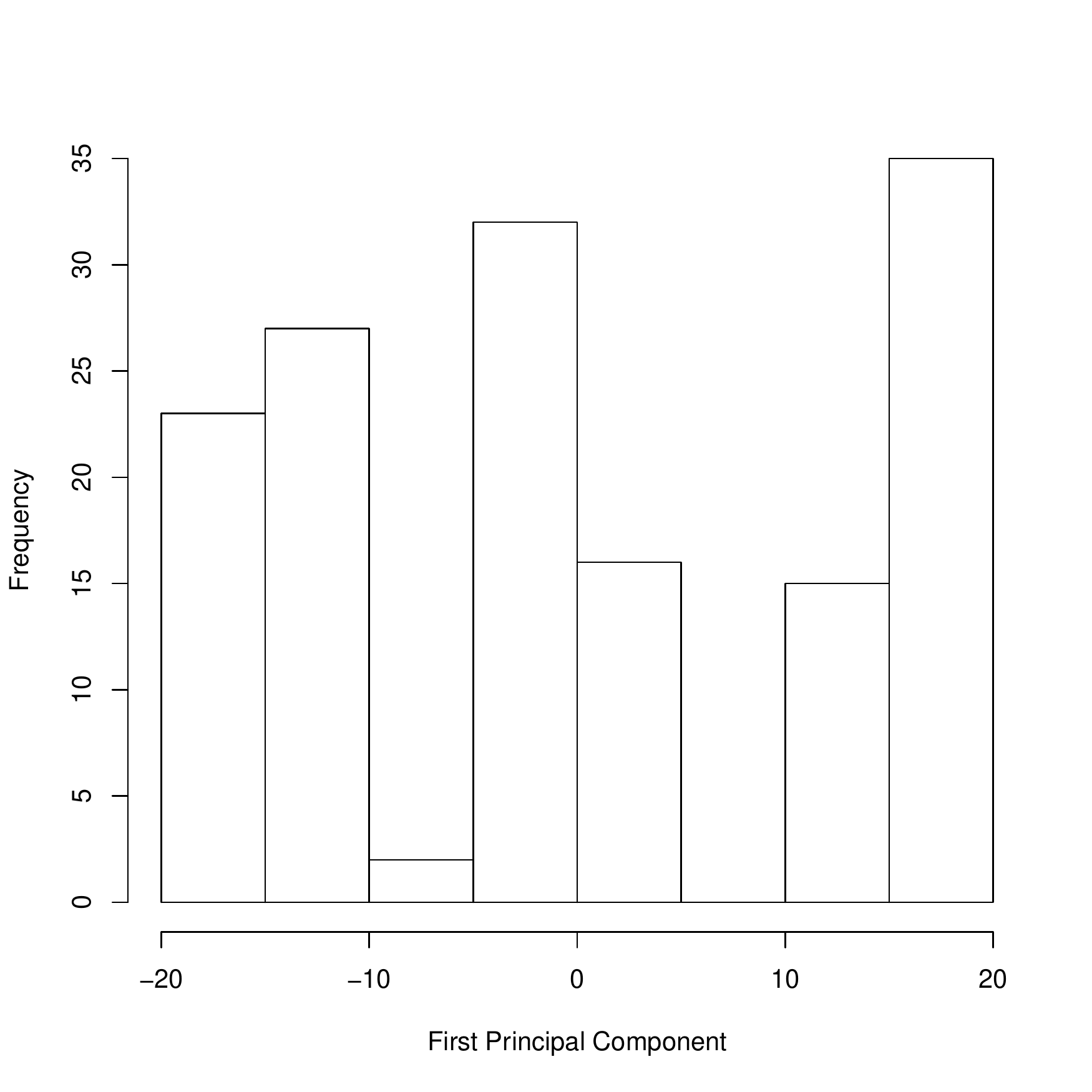}
    \caption{Three Clusters: PCA}
    \label{threeclustpca}
  \end{subfigure}
  \end{center}
  \caption{An illustration of dimension reduction for clusterability evaluation. The left column, Figures
\ref{oneclustoriginal}, \ref{twoclustoriginal} and \ref{threeclustoriginal}, depicts data generated from one, two, or three Gaussian clusters using the same parameters as in Table \ref{simtable1} (row 1) and Table \ref{simtable3} (rows 11 and 16). 
The middle column, Figures \ref{oneclustdist}, \ref{twoclustdist} and \ref{threeclustdist}, includes histograms of the pairwise dissimilarities of the data.  
The right column, Figures \ref{oneclustpca}, \ref{twoclustpca} and \ref{threeclustpca}, includes histograms of the first principal component of the data. 
Distributions of reduced data contained a sole mode for data generated from a single cluster and multiple modes for data generated from multiple clusters.}
\label{visualizedimred}
 \end{figure*}

A large, practical class of clusterability notions rely on one or more of the following: dimensionality reduction and statistical tests of multimodality.
%
The reduced data 
informs the clusterability of the original data, as shown in Figure \ref{visualizedimred}. When data is generated from a single bivariate normal distribution, the original data forms one cluster, and the pairwise distance and first principal component distributions are unimodal. By contrast, when the data is generated from multiple clusters, the pairwise distances and first principal component distributions are multimodal. It is important to note that the the modes of the reduced data need not correspond with the clusters. 
For example, the distribution of pairwise distances generated from three clusters, depicted in Figure \ref{threeclustdist}, has only two modes, corresponding to the smaller within-cluster and larger between-cluster distances. 

In the following two subsections, we briefly review data reduction methods and multimodality tests, before delving into clusterability methods. 

\subsubsection{Data reduction methods}\label{reduction}
Data sets often contain a large number of features, which may 
even outnumber the 
observations. Due to the computational and theoretical challenges associated with high dimensional data, reducing the dimension 
while maintaining the structure of the original data is desirable. 
If dimensions are measured on different scales, then the data may be centered about its mean  or scaled to have unit variance, as recommended in \cite{wold}. 
Specific techniques to reduce data to one dimension are now discussed in greater detail.


One famous data reduction method is principal component analysis (PCA), which 
projects the data onto independent dimensions that explain the original variance \citep{jolliffe2002pca}. There are natural connections between PCA and clustering. 
In fact, 
the principal components (PC) 
correspond to the $k$-means cluster membership indicators \citep{zha2001NIPS,Ding2004}. 
PCA has been recommended to visually inspect for grouping structure \citep{varmuza}.
While multiple components are often retained, the first PC, by definition, explains most of the variation in the data 
\citep{Huber1985,krause2005multimodal}. 
PCA is less prone than other data-reduction methods to the curse of dimensionality but is not well suited to non-linear structures \citep{Huber1985}, for which principal curves \citep{HastieStuetzle,Tibshirani1992}, which produce a non-linear transformation of the data, may be more appropriate. 





The set of dissmilarities between pairs of points in a data set forms another one-dimensional summary.  Distances, {
{which can be calculated using a variety of metrics,}} are often used as 
inputs 
to clustering algorithms, can be calculated for any data set, and have been shown to 
preserve structural features, such as correlation 
\citep{goslee}. 
Yet, distances are sensitive to the curse of dimensionality, potentially yielding misleading results for data with many 
features. Also, 
the use of pairwise distances increases the sample size of the summary to nearly the square of the original size, rendering this approach 
impractical for datasets with a large number of observations. 

\subsubsection{Multimodality tests}\label{multtests}


Intuitively, if a data set contains multiple clusters, then there should be some separation between the clusters. For example, a histogram of pairwise distances should show a group of small distances, representing those within clusters, and a group of large between-cluster distances. On the other hand, homogeneous data should not show such a separation. 
See Figure~\ref{visualizedimred} for an illustration.

A statistical test of multimodality can formally determine if the set of distances for a given dataset has multiple modes, 
indicating that there are multiple clusters. Likewise, tests on data reduced by other methods help detect cluster structure. Multimodality tests \citep{XuEdwardHansonRestrepo} are used for other clustering purposes, such cluster splitting, merging, and validation \citep{dip_means,LuHayesNobelMarron,sigtest,Helgeson,kimes}. 

Multimodality tests initially assume that 
data is generated from a unimodal distribution (the null hypothesis), but may refute that assumption based on the data. The $p$-value is the probability of observing the given input or a more extremely multimodal input 
under the null hypothesis. 
If only a single mode is present, then the $p$-value should be large, indicating that the underlying data is deemed unclusterable. 
By contrast, small $p$-values 
make us question the original assumption of unimodality and instead conclude that multiple modes (and multiple clusters) are present in the population from which the data was generated. 
%

{
{Two multimodality tests are widely used and available in standard software. The dip test 
computes a statistic called the dip, defined as the maximum distance between the empirical distribution and the closest uniform distribution. Details on the dip statistic and the algorithm used to implement it are described elsewhere \cite{hartiganDip,dipprogramjrssc,citediptestR}. 
The dip test rejects the assumption of unimodality if the dip is sufficiently large, indicating that the
data is sufficiently different from the closest 
uniform distribution.  
}} 
Another popular test by 
Silverman 
\cite{silverman} rejects the 
unimodality assumption if the kernel density estimate requires a sufficiently large bandwidth to produce an empirical distribution based on one Gaussian component rather than a Gaussian mixture. {
{In particular, the critical bandwidth ${h}_{crit}$, is defined by Equation \ref{hcrit}
\begin{equation}\label{hcrit}
h_{crit}=\inf\{h: {\hat{f}(.,h)\textnormal{ has at most one mode}}\}
\end{equation}
where
\begin{equation}
\hat{f}(t,h)=n^{-1}h^{-1}\sum_{i=1}^{n}K(h^{-1}(t-X_i)),
\end{equation}
$X_1,\ldots,X_n$ denotes the observed data, and $K$ is the density of the standard normal distribution.
}}



One may be tempted to forgo dimension reduction, apply 
a multimodality test, 
and conclude that the data is clusterable if the data set rejects the null hypothesis for unimodality \citep{NAP1360}.  Unfortunately,  the asymptotic behavior of these multimodality 
tests is unknown when the data is multi-dimensional \citep{NAP1360,HartiganTheory,krause2005multimodal,Wells1978}.
These severe limitations render these methods unpredictable for real data sets, most of which have multiple, if not high dimensions, unless the user first reduces the data to one dimension.

\subsection{Clusterability via Multimodality}\label{multclusttests}
We now introduce previous notions of clusterability 
and several new notions. 
For these notions, if the user fails to reject the null hypothesis of unimodality of the reduced data, then the data does not have clear evidence of cluster structure {\it and should not be clustered}. 
 
 {
 {The majority of this section describes the idea behind each clusterability methods and how it was implemented.  All methods described in this paper were implemented using readily available functions in {\it R} statistical software version 3.3.2 \cite{citeR}. Details are found in Subsections \ref{dipdist}-\ref{PCdip}. 
 
Runtime analysis is presented in Section \ref{runtime}.}}
 This section concludes with a brief summary of notions of clusterability that are not well-suited to the goals of the current analysis.

\noindent 
\subsubsection{
Dip Test on Pairwise Distances (
Dip-dist)}\label{dipdist}
Dip-dist \citep{dip_means} tests for clusters in the set of dissimilarities using the Dip test 
\cite{hartiganDip}. The lengths of the pairwise distances are sufficient for clusterability analysis without needing to consider how the distances are arranged to form the data. Multiple modes in the distance distribution suggest the presence of multiple clusters. {
{The implementation of the dip test is detailed by Hartigan \cite{dipprogramjrssc}. In brief, the (one-dimensional) set of pairwise Euclidean distances is calculated, sorted, and used as inputs to the Dip test. We utilized the {\it dist()} function within {\it R} \cite{citeR} and the {\it dip.test()} function within the {\it diptest} package \cite{citediptestR}.}} 
%

\subsubsection{Silverman Test on Principal Curve (PC Silv.)} 
One method proposes to use Silverman's test 
 for multimodality of the principal curve proposed by Ahmed \cite{Ahmed2012}. The first dimension of the principal curve is extracted and Silverman's test is used to determine if that dimension is unimodal or multimodal. A multimodal principal curve suggests that the original, higher dimensional data exhibits cluster structure.
{
{The implementation of Silverman's test was based on the silvermantest package in {\it R} \cite{citeSilvR}, which includes the calibration recommended by Hall and York \cite{HallYork}. Corrections for both Silverman's original test and the dip test have been proposed \cite{HallYork,ChengHall}, but only the correction for Silverman is available in standard software, such as {\it R}. Principal curves were created using the {\it princurve} package \cite{citeprincurvR}.}}

\subsubsection{Silverman Test on Principal Component (PCA Silv.)}  
A linear alternative to the principal curve 
is to extract the first principal component, also suggested by Ahmed \cite{Ahmed2012}, which explains the maximum variation in the data. 
The Silverman test \cite{silverman,citeSilvR,HallYork} is then applied to this first principal component. {
{The implementation in the paper performs PCA using the singular value decomposition of the centered data, and then the rotated variables are extracted. Computationally, the program calls the {\it prcomp()} function, available in the {\it stats} package in {\it R} \cite{citeR}.}} A multimodal first principal component 
suggests that the original data is clusterable.

\subsubsection{Classic Methods} \label{classic} 
While it is known that multimodality tests may be problematic in higher dimensions, we include these methods in our comparisons for completeness. That is, Classic Silverman (Cl. Silverman) and Classic Dip (Cl. Dip) conduct, respectively, Silverman's and the Dip test of multimodality on the original, multi-dimensional data.

\subsection{Clusterability via Spatial Randomness}
Another method 
\citep{hopkins_original,LawsonJurs1990}, a test of spatial randomness, tells us if a feature is distributed non-randomly across the data set. Hopkins (
Hop.
) 
compares the distances between a sample of data points and their nearest neighbors to the distances from 
a sample of pseudo points  -- with each feature randomly selected from the full data set -- and their nearest neighbors.
If the data are not distributed in clusters, then both sets of 
distances should be similar on average. 
%
Clusterability can be inferred by comparing to a threshold calcuated based on the  distribution of the Hopkins statistic. 
Under the null hypothesis that the data is unclusterable, the test statistic follows a beta distribution with both parameters equal to the number of points selected to sample $n$ \citep{hopkins_original,LawsonJurs1990}. 
Thus, Hopkins' statistic should be compared to a Beta quantile $q_\alpha(n,n)$. 
{
{The Beta quantile $q_\alpha(n,n)$ is defined as the value such that, assuming the data was generated without clusters, the chance of concluding that the data is clustered, i.e. $P(H<q_\alpha(n,n))$ is $100\alpha\%$. We use a one-sided test, because if the data were more spatially random than expected by chance, 
it would still be considered unclusterable.}} 
%
Yet, the choice of $n$ requires caution. According to Lawson and Jures \cite{LawsonJurs1990}, ``if too few points are chosen, then the nearest-neighbor distances chosen will not be representative of the entire distribution of distances.'' If too many points are chosen, 
\cite{Dubes1987} warn that the ``assumptions about the Beta distribution will be invalid.'' Previous authors recommend sampling $5-10\%$ of the data \citep{LawsonJurs1990,banerjeeDave}. In this paper, we use
a 10\% sampling rate. {
{In this manuscript, Hopkins was implemented using the {\it hopkins()} function in the {\it clustertend} package \cite{citeHopkinsR}.}}


%
%

\subsection{New Clusterability Methods}\label{ournewmethods}
This section describes our new proposed approaches for evaluating clusterability. 
Since both the dip and Silverman's test are valid multimodality tests, 
we propose to use both on each reduced version of the data. To our knowledge, the following methods below have not been previously proposed. 

\subsubsection{Silverman's test on dissimilarities (Silv.-dist)} 
Rather than using the dip test on the set of pairwise Euclidean distances \citep{dip_means}, we propose to use Silverman's test, with the necessary correction 
\cite{HallYork}.

\subsubsection{Dip test on principal component (PCA Dip)}
Instead of using Silverman's test of whether the first principal component is multimodal, this method uses the dip test.

\subsubsection{Dip test on principal curve (PC Dip)}\label{PCdip} 
The dip test classifies the modality of the principal curve. 

\subsection{Other Clusterability Methods}
Some notions of clusterability in the literature have been omitted from our study, as they are either impractical or otherwise unsuited to our goals. 
The effectiveness requirement allowed us to eliminate several notions of clusterability. For example, worst pair ratio~\citep{Epter} identifies data as clusterable if and only if there is a $k$-clustering where the minimum between-cluster distance is greater than the maximum in-cluster distance. Despite the simplicity and elegance of this notions, it identifies many clearly clusterable data sets as unclusterable (for example, three Gaussian distribute clusters positioned closer to each other, so that the separation between them is smaller than the radius of the clusters), and as such fails the effectiveness condition. 


Other elegant clusterability measures that have been useful in theoretical analysis are omitted from our study since they are either NP-hard to compute \citep{ackerman2009clusterability} or too strict for practical application due to their high sensitivity to noise and outliers. Since we 
seek notions that are efficient, 
and applicable in practice, we had to omit all such measures from our analysis. Notions based on specific algorithms or objective functions~\citep{ ackerman2009clusterability, awasthi2010stability, awasthi2012center, balcan2009approximate, LuHayesNobelMarron,Swamy}, 
are omitted from our 
analysis, which 
seeks to identify the presence of any cluster structure, not only that which can be discovered by a specific clustering technique. Approaches to clusterability, such as \cite{Szczubialka}, relying on subjective judgment rather than a quantifiable measure, are also omitted.

\section{Simulations}\label{sims}

Our analysis of clusterability measures begins with extensive simulations. The purpose of the simulations is twofold. First, simulations allows us to perform basic tests to ensure that clusterability measures behave reasonable on clearly clusterable data and clearly unclusterable data. If we discover, for instance, that a measure fails to identify data generated from two well-separated Gaussian distrbutions as clusterable, then the measure fails our effectiveness requirement. Similarly, failure to identify data generated from a single Gaussian distribution as unclusterable would preclude the application of a measure in practice, as it would contradict basic intuition about what it means to be clusterable. 

The other major goal achieved by simulations is identifying under what conditions different measures are most appropriate. In addition to testing clear cut cases, we also include simulations with noise, varied number of cluster sizes and diameters, outliers, etc, to check for robustness of clusterability methods.  For example, in some contexts, the inclusion of a single distant outlier in otherwise unclusterable data may indicate the formation of a second cluster. Yet, for other applications, the outlier is best ignored. Our simulations let us identify which clusterability technique allows for small clusters, and which exhibit outlier robustness. This information can then be used by clustering users to make informed choices when selecting a clusterability approach for their data. 

{
{One of the main approaches to the study of clustering involves the analysis of common statistical distributions (see, for example, \cite{NordhaugMyhre2018491, celeux1995gaussian, dasgupta1999learning})}.}
Our extensive simulations evaluate each approach to clusterability using all clusterability tests in Sections \ref{multclusttests}-\ref{ournewmethods}. 
The simulations include 31 types of data sets, each generated with the same parameters 1000 times, for a total of 31,000 simulations. Simulations consist of clusters generated from one or more Gaussian or t-distributions, sometimes with a small number of outliers, and chaining data with one or more lines or circles. 
Simulations were performed in {\it R} version {{3.3.2}} \cite{citeR}. Code is found at the following link: \url{http://www.mayaackerman.info/clusterability.R}. Further details for each simulation scenario are included in Section \ref{simdetails}.

{
\begin{table*}
 \centering
  \begin{tabular}{{rcrllr}}
    \hline
Row&Clusters&Dim&Distribution(s)&Clust size&Total\\ \hline
1&1&2&$N(100,2)$&50&50\\
\hline
2&1&3&$N(100,2)$&50&50\\
\hline
3&1&10&$N(2,2)$&50&50\\
\hline
4&1&50&$N(2,2)$&100&100\\
\hline
5&1&2&$N(50,2)$&50&51\\
&+1 outlier&&$N(\mu,2): \mu\sim U(60,65)$&1\\ \hline
6&1&2&$N(50,2)$&250&251\\
&+1 outlier&&$N(\mu,2): \mu\sim U(60,65)$&1\\ \hline
7&1&2&$N(50,2)$,&50&53\\ 
&+3 outliers&&$N(\mu,2):$ $\mu\sim U((40,55),(45,60))$&1\\ 
&&&$N(\mu,2):$ $\mu\sim U((65,65),(70,70))$&1\\
&&&$N(\mu,2):$ $\mu\sim U((65,45),(70,50))$&1\\
\hline
8&1&2&$T_5(100)$&100&100\\ \hline
9&1&2&$T_{10}(100)$&100&100\\ \hline
10&1&2&$T_{15}(100)$&100&100\\ \hline
11&2&2&$N(30,2)$, $N(50,2)$&50&100\\ \hline
12&3&2&$N((30,20),2)$,  $N((40,20),2)$,  $N((35,30),2)$&50&150\\ \hline
13&3&2&$N((30,40),2)$, $N((70,40),2)$,  $N((50,80),2)$&50\\ 
&+noise&&$N(50,20)$&80&230\\ \hline
14&3&2&$N((30,20),1)$,  $N((40,20),3)$,  $N((35,30),5)$&50&150\\ \hline
15&3&2&$N((35,40),2)$,  $N((65,40),2)$,  $N((50,60),2)$&100,66,33&199\\ \hline
16&3&2&$N((35,40),2)$,  $N((65,40),2)$,  $N((50,60),2)$&50&150\\     \hline
17&3&2&$N(20,2)$, $N(40,2)$,  $N(60,2)$&50&150\\     \hline
18&2&10&$N(10,2), N(20,2)$&50&100
\\ \hline
19&4&10&$N(10,2), N(20,2), N(60,2), N(80,2)$&50&200
\\ \hline
20&2&50&$N(5,2)$,  $N(10,2)$&100&200\\ \hline
21&2&50&$N(3,2)$,  $N(6,2)$&100&200\\ \hline 22&2&2&$T_5(50)$, $T_5(150)$&100&200\\ \hline
23&2&2&$T_{10}(50)$, $T_{10}(150)$&100&200\\ \hline
24&2&2&$T_{15}(50)$, $T_{15}(150)$&100&200\\ \hline
25&1&2&unit circle&50&50\\ \hline
26&2&2&2 concentric circles (radii: 1,2) &50&100\\ \hline
27&3&2&3 concentric circles (radii: 1,2,3) &50&150\\ \hline
28&5&2&5 concentric circles (radii: 1,2,3,4,5) &50&250\\ \hline 
29&1&2&1 line ($x=50$, $y\sim N(50,25)$) &100&100\\ \hline
30&2&2&2 lines ($x_1=30$, $x_2=55$, $y\sim N(50,25)$) &100&200\\ \hline
31&2&2&circle (radii: 3) at origin &100&200\\ 
&&&line ($x=5$, $y\sim N(0,2)$)&100\\ \hline
\end{tabular}
  \caption{Parameters Used in Simulations. 
Simulations include clusters with Gaussian, T, and uniform distributions, separate outlying points, as well as parallel lines and circles. Dimensions are generated independently and identically unless specified otherwise. Notation is described in Section \ref{simdetails}.} 
   \label{simparms}
\end{table*}
}



Tables \ref{simtable1} through \ref{simtable4} include the 
the percentage of data sets on which 
the test yielded a $p$-value less than $0.05$, indicating that the tests rejected the null hypothesis of unimodality at the traditionally used 5\% significance level. \emph{High values in Tables \ref{simtable1}-\ref{simtable4}  indicate clusterable data, while low values indicate poor clusterability. }
For unambiguously unclusterable data sets, the proportion of rejections corresponds to type I error, the rate of erroneously classifying data sets generated without clusters as clusterable. Type I error greatly exceeding 5\% indicates that the method is invalid and produces excessive false positives.

For unambiguously clusterable data sets, the proportion of rejections corresponds to the statistical power, or ability of the test to correctly classify clusterable sets as having cluster structure. Higher power is desirable. 
Yet, results are complicated by the ambiguous nature of clustering. When a small number of outlying points are present, the decision to classify the data as clusterable depends on whether outliers should be considered as small clusters. 

\subsection{Simulation Details}\label{simdetails}
{
{Table \ref{simparms} summarizes the parameters used in the simulations. Unless otherwise specified, observations have independent and identically distributed dimensions with the given parameters. $N(\mu,\sigma)$ denotes data generated from independent Gaussian distributions in each dimension with mean $\mu$ and standard deviation $\sigma$. For example, lines 3 and 4 each include data generated from a single Gaussian cluster in 10 and 50 dimensions, respectively, each with mean and standard deviation equal to 2.  $T_{d}(\lambda)$ denotes data generated independently in each dimension from a $T$ distribution with $d$ degrees of freedom and non-centrality parameter $\lambda$. In lines 22-24, each scenario was generated from two 2-dimensional non-central $T$ distributed clusters with either 5,10, or 15 degrees of freedom and non-centrality parameters of 50 and 150. $U(a,b)$ means that data independently generated in each dimension from a uniform distribution with density equally distributed  between $a$ and $b$. 

When dimensions are non-identical, each dimension is specified as needed. For example, $N((\mu_1,\mu_2,\ldots,\mu_d),\sigma)$ denotes data generated from a $d-$dimensional Gaussian distribution with means located at $\mu_i$ for independent dimensions $i=1,2,\ldots,d$ and standard deviation $\sigma$ in each dimension. In particular, in line 12, data was generated from three two-dimensional Gaussian distributions, each with standard deviation 2 in each independent dimension, and centers at $(30,20)$,   $(40,20)$,  and $(35,30)$.} 

In some cases, additional points are added to represent outliers and noise. Line 13 describes data generated from three two-dimensional Gaussian distributions, similar to line 12 but with means at $(30,40)$,  $(70,40)$,  and $(50,80)$, and a fourth, much wider, Gaussian distribution with mean 50 and standard deviation 20 in both dimensions. The outliers in lines 5-7 include between one and three additional Gaussian observations generated with high probability to be far apart from the rest of the data. In particular, while all two-dimensional data in line 5 is generated from a Gaussian distribution with standard deviation 2 in each dimension, the main cluster of 50 points was generated from a mean at (50,50), and the remaining point was generated with mean in each dimension randomly selected from a uniform distribution bounded by 60 and 65.

A subset of simulations, shown in lines 25-31 of Table \ref{simparms} include data generated with chaining structure, including one or more circles or lines. The circles were all centered at the origin (0,0), and the radius of each was an integer between 1 and 5. The lines were each vertical, with constant x-coordinates (denoted in the table by $x=c$) and y-coordinate generated from a univariate Gaussian distribution. Line 31, which includes both a circle and a lines exemplifies this notation. The circle, centered at (0,0) has a radius of 3, and the line, located to the right of the circle at $x=5$, has points vertically generated from a Gaussian distribution with mean 0 and standard deviation 2.}

\subsection{Type I error: Results for Unclusterable Data}
Principal curve methods were invalid, concluding that single-cluster data sets were clusterable at a much higher rate than 5\%.
Hopkins, PCA-Silverman and classic Silverman have type I error around 5\% as expected in two dimensions.
\footnote{For valid methods, values in Table 1 should be below or reasonably close to 0.05. If the true false positive rate is 5\%, then we would expect with 95\% confidence that the observed value should be below $0.05+1.96*\sqrt{0.05*0.95/1000}\approx0.064$.
 Based on this threshold, PCA Silverman has slightly inflated type I error in 50 dimensions, and Classic Silverman has inflated type I error in 3 dimensions. However, because we would expect 5\% of the results for unclusterable data sets to exceed this value, it is not unusual to see 2 results with slightly inflated type I error rates. In fact, if we adjust for the total number of comparisons for unclusterable data, then the false positive rates would be compared to a different threshold (0.072) and would not be considered excessive. 
} 
Hopkins (in multiple dimensions) and distance based methods have excessively low type 1 error of less than 1\%, indicating that they may be overly conservative. 
All methods except principal curves have 
low false positive rates for single Gaussian clusters. 


\subsection{Performance with Outliers and Small Clusters}
When 
outlying points are introduced to otherwise unclusterable data, one could argue either for or against clusterability.  Methods vary in their conclusions: Dip-based methods classify the data as unclusterable, while the Hopkins statistic and Silverman-based methods classify such data as clusterable, identifying the outliers as separate clusters.  
Dip-based methods consider the data clusterable less than 10\% of the time, even for t-distributions with 5 degrees of freedom, when multiple outliers are likely; Hopkins and Silverman-based methods frequently conclude that the data is clusterable, ranging from 44\% to 85\% of the time. As expected, the proportion decreases as the degrees of freedom increases and the distribution converges to Gaussian.  
%
\emph{Where the dip test is robust to outliers, Silverman's test and the Hopkins statistic allow for small clusters}. This finding reflects the inherent ambiguity of clustering; for some applications, small clusters are acceptable, while for others, robustness to outliers is desired. In fact, clustering algorithms display the same phenomenon: 
some tend to identify small clusters, while others effectively view such data as outliers ~\citep{oligarchies}. 

\begin{table*}
 \centering
  \begin{tabular}{{llccccccccc}}
    \hline
    & & \textbf{Dip} & \textbf{Silv} & \multirow{2}{*}{\textbf{Hop}}   & \textbf{Cl.} & \textbf{Cl.} & \textbf{PCA} & \textbf{PCA} & \textbf{PC}	&\textbf{PC} \\
& {\textbf{Data}}& \textbf{Dist} & \textbf{Dist} &    & \textbf{Dip} & \textbf{Silv} & \textbf{Dip} & \textbf{Silv} &\textbf{Dip}	& \textbf{Silv} \\
    \hline
    \emph{1.} & 1 cluster 2D & 0.000 & 0.042 & 0.057 & 0.001 & 0.055 & 0.001 & 0.053 &{\sout{0.179}} & {\sout{0.346}} \\
    \hline
    \emph{2.} & 1 cluster 3D & 0.000 & 0.042 & 0.012 & 0.000 & 0.068 & 0.002 & 0.062 & {\sout{0.213}} & {\sout{0.444}}\\
    \hline
    \emph{3.} & 1 cluster 10D & 0.000 & 0.035 & 0.000 & 0.000 & 0.055 & 0.003 & 0.057 & {\sout{0.235}} & {\sout{0.585}}\\
    \hline
    \emph{4.} & 1 cluster 50D & 0.000 & 0.033 & 0.000 & 0.000 & 0.052 & 0.002 & 0.064 & {\sout{0.220}} & {\sout{0.764}}\\
    \hline
  \end{tabular}
   \caption{Results of Simulations for Data Generated from A Single Cluster. 
Proportion of data sets considered clusterable out of the 1000 data sets for each type. Entries are interpreted as type I error, which should not greatly exceed 5\%. Strike-through denotes entries that have excessive Type I error. Methods had reasonably low false positive rates, with the exception of those using principal curves. 
}
   \label{simtable1}
\end{table*}

\begin{table*}
 \centering
  \begin{tabular}{{llccccccccc}}
  
  \hline
    & & \textbf{Dip} & \textbf{Silv} & \multirow{2}{*}{\textbf{Hop}}   & \textbf{Cl.} & \textbf{Cl.} & \textbf{PCA} & \textbf{PCA} & \textbf{PC}	& \textbf{PC} \\
& {\textbf{Data}}& \textbf{Dist} & \textbf{Dist} &    & \textbf{Dip} & \textbf{Silv} & \textbf{Dip} & \textbf{Silv} & \textbf{Dip}	& \textbf{Silv} \\
    \hline
    \emph{5.} & 1 cluster 2D \\
    &with outlier & {0.000} & $0.987$ & 0.858 & {0.000} & 0.890  & {0.005} & $0.998$ & {{0.136}} & ${0.984}$\\
    \hline
    \emph{6.} & 1 large cluster 2D \\
    &with outlier & {0.000} & $0.975$ & $1.000$ & {0.000} & $0.911$ & {0.001} & $0.989$ & {{0.331}} & {{0.980}}\\
    \hline
    \emph{7.} & 1 cluster 2D \\
    &with 3 outliers & 0.101 & $0.976$ & 0.815 & {0.000} & $0.954$ & {0.003} & $0.942$ & {0.016} & {0.935}\\
    \hline
    \emph{8.} & 1 T-dist cluster \\
    &with  df=5 & {0.007} & 0.573 & 0.852 & {0.000} & 0.440 & {0.000} & 0.463 & {0.070} & 0.490\\
    \hline
    \emph{9.} & 1 T-dist cluster \\
    &with df=10 & {0.000} & 0.214 & 0.657 & {0.000} & 0.240 & {0.000} & 0.282 & {0.095} & 0.348\\
    \hline
    \emph{10.} & 1 T-dist cluster \\
    &with df=15 & {0.000} & 0.117 & 0.579 & {0.002} & 0.209 & {0.000} & 0.220 & {0.098} & 0.344\\
    \hline
  \end{tabular}
   \caption{Results of Simulations Generated from Data with Outliers. 
Proportion of data sets considered clusterable out of the 1000 data sets of each type. Methods differed in their conclusions on such data sets, reflecting the ambiguous nature of clustering for data with outliers. Methods using Hopkins or Silverman tended to consider outliers as separate clusters, while dip-based methods tended to be robust to outliers. 
} 
   \label{simtable2}
\end{table*}

\begin{table*}
 \centering
  \begin{tabular}{{llccccccccc}}
    \hline
    & & \textbf{Dip} & \textbf{Silv} & \multirow{2}{*}{\textbf{Hop}}   & \textbf{Cl.} & \textbf{Cl.} & \textbf{PCA} & \textbf{PCA} & \textbf{PC}	& \textbf{PC} \\
& {\textbf{Data}}& \textbf{Dist} & \textbf{Dist} &    & \textbf{Dip} & \textbf{Silv} & \textbf{Dip} & \textbf{Silv} & \textbf{Dip}	& \textbf{Silv} \\
    \hline   
    \emph{11.} & 2 separated \\
    &clusters 2D & 1.000 & 1.000 & 1.000 & 1.000 & 1.000 & 1.000 & 1.000 & 1.000 & 1.000\\
    \hline
    \emph{12.} & 3 close \\
    &clusters 2D & 0.996 & 1.000 & 0.789 & 1.000 & 1.000 & 0.801 & 0.974 & 0.757 & 0.826\\
    \hline
    \emph{13.} & 3 noisy \\
    &clusters 2D & 1.000 & 0.996 & 0.989 & 0.999 & 0.995 & 1.000 & 0.999 & 1.000 & 0.983\\
    \hline
    \emph{14.} & 3 clusters 2D, \\
    &varied radii & 1.000 & 1.000 & 1.000 & 1.000 & 1.000 & 1.000 & 1.000 & 1.000 & 1.000\\
    \hline
    \emph{15.} & 3 clusters 2D, \\
    &varied density & 1.000 & 1.000 & 1.000 & 1.000 & 1.000 & 1.000 & 1.000 & 1.000 & 1.000\\
    \hline
    \emph{16.} & 3 separated \\
    &clusters 2D & 1.000	 & 1.000 & 1.000 & 1.000 & 1.000 & 1.000 & 1.000 & 1.000 & 1.000\\
    \hline
    \emph{17.} & 3 separated \\
    &clusters 3D &1.000 & 1.000 & 1.000 & 1.000 & 1.000 & 1.000 & 1.000 & 1.000 & 1.000\\
    \hline
    \emph{18.} & 2 separated \\
    &clusters 10D & 1.000 & 1.000 & 1.000 & 1.000 & 1.000 & 1.000 & 1.000 & 1.000 & 1.000 \\
    \hline
    \emph{19.} & 4 separated \\
    &clusters 10D & 1.000 & 1.000 & 1.000 & 1.000 & 1.000 & 1.000 & 1.000 & 1.000 & 0.884\\
    \hline
    \emph{20.} & 2 close \\
    &clusters 50D & 1.000 & 1.000 & 1.000 & 0.691 & 0.999 & 1.000 & 1.000 & 1.000 & 1.000\\
    \hline
    \emph{21.} & 2 partially \\
    &overlapping 50D & 1.000 & 1.000 & \bf{0.445} & \bf{0.000} & \bf{0.041} & 1.000 & 1.000 & 0.995 & 0.997\\
    \hline
	\emph{22.} & 2 T-dist cluster \\
    &with df=5 & 1.000 & 1.000 & 1.000 & 1.000 & 1.000 & 1.000 & 1.000 & 1.000 & 1.000\\
    \hline
    \emph{23.} & 2 T-dist cluster \\
    &with df=10 & 1.000 & 1.000 & 0.999 & 1.000 & 0.998 & 1.000 & 1.000 & 1.000 & 0.999\\
    \hline
    \emph{24.} & 2 T-dist cluster \\
    &with df=15 & 1.000 & 1.000 & 1.000 & 1.000 & 1.000 & 1.000 & 1.000 & 1.000 & 1.000\\
    \hline
  \end{tabular}
   \caption{Results of Simulations Generated with Multiple Clusters. 
Proportion of data sets considered clusterable out of the 1000 data sets of each type. These numbers corresond to statistical power to detect the cluster structure. Numbers in bold denote methods with much lower power than others. In particular, Hopkins and especially classic multimodality tests lose power in high dimensions. All other methods had reasonably high power for these simulations.
\\
} 
   \label{simtable3}
\end{table*}

\begin{table*}
 \centering
  \begin{tabular}{{llccccccccc}}
    \hline
    & & \textbf{Dip} & \textbf{Silv} & \multirow{2}{*}{\textbf{Hop}}   & \textbf{Cl.} & \textbf{Cl.} & \textbf{PCA} & \textbf{PCA} & \textbf{PC}	& \textbf{PC} \\
& {\textbf{Data}}& \textbf{Dist} & \textbf{Dist} &    & \textbf{Dip} & \textbf{Silv} & \textbf{Dip} & \textbf{Silv} & \textbf{Dip}	& \textbf{Silv} \\
    \hline
 	\emph{25.} & Single\\
    &Circle & 0.010 & {\sout{0.309}} & {\sout{0.837}} & {\sout{0.951}} & {\sout{0.945}} & {\sout{0.909}} & {\sout{0.988}} & {\sout{0.626}} & {\sout{0.775}}\\
    \hline
    \emph{26.} & 2 concentric \\
    &circles & 1.000 & 1.000 & 0.873 & 0.533 & 0.751 & 0.322 & 0.472 & 0.619 & 0.802\\
    \hline
    \emph{27.} & 3 concentric \\
    &circles & 1.000 & 1.000 & 0.894 & 0.167 & 0.486 & 0.079 & 0.193 & 0.607 & 0.774\\
    \hline
    \emph{28.} & 5 concentric \\
    &circles &1.000 & 1.000 & 1.000 & 0.364 & 0.394 & 0.159 & 0.364 & 	0.726 & 0.895\\
    \hline
 	\emph{29.} & Single\\
    &Line & 0.004 & 0.049 & {\sout{0.378}} & 0.000 & {\sout{0.112}} & 0.000 & 0.055 & {\sout{0.642}} & {\sout{N/A}}\\
	\hline
	\emph{30.} & 2 parallel \\
    &lines & 1.000 & 0.889 & 1.000 & 1.000 & 0.996 & 0.000 & 0.055 & 0.997 & 0.989 \\
    \hline
 	\emph{31.} & Line and\\
    &Circle & 0.998 & 0.999 & 1.000 & 1.000 & 0.958 & 0.209 & 0.894 & 0.876 & 0.982\\
    \hline\\
  \end{tabular}
   \caption{Results of Simulations for Chaining Data. 
Proportion of data sets considered clusterable out of the 1000 data sets of each type. Results correspond to type I error for data sets with a single cluster circle or line and statistical power for data sets with multiple circles and/or lines. Strike-through denotes entries that have excessive Type I error. Looking at the columns shows that the only method that had type I error controlled, and thus the only valid method for chaining data, is Dip-Dist.
Principal curves, which are non-linear reductions, 
failed to converge for linear data. 
\\
} 
   \label{simtable4}
\end{table*}

\begin{table*}
  \centering
  \begin{tabular}{{lrrrrrrrrrrr}}
    \hline
     &&& \textbf{Dip} & \textbf{Silv} & \multirow{2}{*}{\textbf{Hop}}   & \textbf{Cl.} & \textbf{Cl.} & \textbf{PCA} & \textbf{PCA} & \textbf{PC}	& \textbf{PC} \\
 {\textbf{Data}}&{\textbf{n}}&{\textbf{d}}& \textbf{Dist} & \textbf{Dist} &    & \textbf{Dip} & \textbf{Silv} & \textbf{Dip} & \textbf{Silv} & \textbf{Dip}	& \textbf{Silv} \\
    \hline
    Faithful&272&2& $0.0000$ & $0.0000$ & 1.00 & $0.0000$ & $0.0000$ & 0.0017 & $0.0000$&$0.0000$&$0.0000$\\
    \hline
    Iris& 150&4& $0.0000$ & $0.0000$ & 1.00 & 0.0014 & $0.0010$ & $0.0000$ & $0.0000$&0.0164&0.0022 \\
    \hline
    \hline
    Rivers&141&1 & 0.2772 & $0.0000$ & 0.92 & 0.9922 & 0.0192 & 0.9922 & 0.0334&0.9922&0.0291\\
    \hline
    Swiss &47&6& $0.0000$ & $0.0000$ & 0.41 & 0.1386 & $0.0000$ & 0.0001 & $0.0000$&$0.0000$&0.0010\\
    \hline
  \end{tabular}
  \caption{Results of Clusterability Tests for Non-Simulated Data with Previously Known or Visually Ambiguous Cluster Structure. 
This table presents the $p$-values, rounded to the nearest ten-thousandth,  for the each clusterability test on real data sets from the R Datasets package. Recall that $p<0.05$ signals clusterable data and $p\geq 0.05$ signals that data is unclusterable at the 5\% significance level. The Hopkins value presented is the proportion of the time out of 100 runs that the Hopkins statistic was below the appropriate beta quantile. For the Hopkins results, high values indicate clusterability. 
The top two rows illustrate famous datasets known to possess multiple large clusters. The bottom two rows include datasets with a small number of outliers. The letter ``n'' refers to the number of observations in the dataset, and ``d'' the number of dimensions, or features. }
\end{table*}

\begin{table*}
  \centering
  \begin{tabular}{{lrrrrrllllll}}
    \hline
     &&& \textbf{Dip} & \textbf{Silv} & \multirow{2}{*}{\textbf{Hop}}   & \textbf{Cl.} & \textbf{Cl.} & \textbf{PCA} & \textbf{PCA} & \textbf{PC}	& \textbf{PC} \\
 {\textbf{Data}}& {\textbf{n}}&{\textbf{d}}&\textbf{Dist} & \textbf{Dist} &    & \textbf{Dip} & \textbf{Silv} & \textbf{Dip} & \textbf{Silv} & \textbf{Dip}	& \textbf{Silv} \\
    \hline
    Attitude&30&7 & 0.9040 & 0.9598 & 0.00  & 0.9113 & 0.9150 & 0.6846 & 0.1534&0.1823&0.2174\\
    \hline
    Cars &50&2& 0.6604 & 0.9931 & 0.19  & 0.8613 & 0.3396 & 0.8320 & 0.4213&0.7680&0.5866\\
    \hline
    Trees&31&3 & 0.3460 & 0.2900 & 0.18 & \bf{0.0001} & $\bf{0.0000}$ & 0.8414 & 0.3675&0.6717&0.2282\\ 
    \hline
    Ratings &43&12& 0.9938 & 0.7313 & 0.69 & \bf{0.0014} & \bf{0.0187} & 0.8550 & 0.1412&0.4501 &\bf{0.0000} \\
    \hline
    Arrests &50&4& 0.9394 & 0.1887 & 0.01 & 0.6261 & \bf{0.0171} & 0.5545 & 0.1286 & \bf{0.0045}&$\bf{0.0000}$\\
    \hline
  \end{tabular}
  \caption{Results of Clusterability Tests for Non-Smulated Data Lacking Known Cluster Structure. 
This table presents the $p$-values, rounded to the nearest ten-thousandth, for the each clusterability test on real data sets from the R Datasets package. The Hopkins value presented is the proportion of the time out of 100 runs that the Hopkins statistic was below the appropriate beta quantile. For the Hopkins results, high values indicate clusterability.  Recall that $p<0.05$ signals clusterable data (entries are in {\bf bold}) and $p\geq 0.05$ signals that data is unclusterable  at the 5\% significance level. Notably, the only bold entries are from methods that performed poorly in simulations.
}
\end{table*}


\subsection{Power: Results for Clusterable Data}
When clusters were well-separated, all methods approached or reached 100\% power, even in the presence of noise. 
When two fifty-dimensional clusters were close to each other {
{as in row (20)}}, all methods have nearly perfect power except for Classic Dip 
with around 70\% power. For partially overlapping 50D clusters, e.g. row (21), the power of the Hopkins test drops to 32\% and both classic methods drop below 5\%. {\it This indicates that classical methods perform poorly in high dimensions for overlapping clusters.} By contrast, utilizing either PCA or pairwise distances, 
both Dip and Silverman tests maintain near perfect power to detect the presence of close or overlapping high dimensional clusters.
We also examine 
two-dimensional data generated from independent t-distributions with 5, 10, and 15 degrees of freedom. 
All methods have nearly 100\% power 
to detect the t-distributed 
clusters.
Most methods had high power 
to detect three or four clusters, except that power for PCA dip and the Hopkins statistic dropped when the separation between clusters decreased. 
Specifically, classic, distance-based methods, 
and PCA-based methods considered all or nearly all (95+\%) datasets as clusterable. 

\subsection{Results for Data with Chaining Structure}
Finally, we examine data with chaining structure, including a single line, two parallel lines, one, two, three, and five concentric circles, and both a line and a circle. For data arranged in one line, classic dip, PCA methods and distance methods did not conclude that the data had multiple clusters. Hopkins classified the line as clusterable nearly 40\% of the time, and classic Silverman concluded the data had structure over 10\% of the time. Surprisingly, all methods except dist-dip considered a single circle as clusterable. Distance Silverman concluded that the single circle had cluster structure about 30\% of the time, while PCA, classical methods, and Hopkins concluded the same over 85\% of the time. Principal curve methods nearly always failed to converge for data comprising a single line. \emph{Thus, dip-dist may be the only valid method for chaining data.}

While both classic and PCA methods have relatively low power to detect the inherent structure of multiple groups of chaining data,  distance-based methods and Hopkins continue to detect the clusters. 
Multiple parallel lines,  depicted in 
row 18, are considered  clusterable by  distance based methods and reasonably well (87\% power) by Hopkins. 
PCA based methods have less than {{6\% power
}}, failing to detect the separate lines most of the time. 
Distance based methods have 100\% power and Hopkins has nearly 90\% power to detect distinct circles, while PCA and classic methods have reduced power for 2 or 3 circles. {
{Curiously, power for PCA methods did not change monotonically with the number of circles. Power was highest for two circles, lowest for three circles, and in between for five circles. However, strange behavior for PCA based methods is likely due to fact that PCA forms linear projections, which may not be appropriate for clearly non-linear data, such as circles.}}

All methods had high power ($\geq$89\%) to detect cluster structure in data consisting of one circle and one line except PCA dip, which only concluded the data was clusterable 20\% of the time. \emph{In sum, dip-dist was the most effective method for chaining data, retaining high power to detect clustered chaining data and being the only method that didn't excessively conclude that data generated to lack groups was clusterable.}

\section{Results on Non-Simulated Data}

%
%
In this section, we apply our methods of clusterability evaluation to non-simulated data sets from the {\it R} \emph{datasets} package, {
{a collection of datasets from prior studies}}.\footnote{Due to the use of sampling in Hopkins' method, we run the method 100 times for each dataset and report the proportion of $p$-values less than 0.05.} 
The datasets we present 
were selected {
{among data sets of standard structure (e.g. time series were not considered)}} to ensure sufficient sample size 
and varied dimension. 
For the sake of completeness, we include all 
tests, but the reader should recall based on Section \ref{sims} that some tests may be inappropriate under various conditions (e.g. principal curve methods due to their inflated false positive rates and classic tests for data with multiple dimensions).
References for all data sets \citep{fisher1936use,hardle1991smoothing, azzalini1990look, AlvarezPena,mcneil1977interactiveprimer,mosteller1977data,ezekiel1930methods,chatterjee1991regression,ryan1976minitab,Register} were examined for evidence of previously known cluster structure. Data sets were also visually inspected for clear outliers, which could indicate ambiguous structure. {
{Figures in an accompanying article depict two-dimensional projections of each dataset and histograms of the data reduced by distances and PCA, respectively. All methods were implemented using {\it R} functions as described in section \ref{multclusttests}.}}
%

Overall, results of the clusterability tests were consistent with expectations based on the references and simulations, detailed in Section \ref{sims}. 
Methods agreed in capturing clear structure in famous data sets and differed in their treatment of data with ambiguous structure, showing similar response to outliers as in the simulations. (The use of Silverman's tests corresponded to considering outliers as separate clusters, while the use of the dip test corresponded to considering outliers as noise.) Similarly, methods generally failed to declare cluster structure in data sets with no prior evidence of clusters, with a few expections. For example, the finding for classic methods to deem some multidimensional data sets as clusterable when no prior evidence could be found to confirm the existence of clusters was consistent with prior knowledge that the methods have unknown asymptotic behavior. Principal curve methods both considered as clusterable some data sets that lacked known evidence of cluster structure and failed to converge on a famous highly linear data set, paralleling its problems in simulations with convergence and excessive false positives. 
Specific results for each dataset follow.


Two famous data sets that were known {\it a priori} to have cluster structure were considered clusterable under all methods. First, the 
\emph{iris} data set~\citep{fisher1936use} 
is 
known to have three clusters corresponding to three species of iris flowers. 
%
%
%
%
Second, the \emph{faithful} data set~\citep{hardle1991smoothing, azzalini1990look}, which captures eruption duration and waiting time 
for the Old Faithful geyser, 
has previously been shown to have two groups \citep{AlvarezPena}. 
All of the tests 
 conclude that both data sets are clusterable, agreeing with  previous knowledge. 


Paralleling our simulations, 
we find that methods relying on the Hopkins statistics or the Silverman tests may be preferred when small clusters are of interest, while techniques using the Dip test may be desired when the application calls for  robustness to outliers. The one-dimensional \emph{rivers} data set~\citep{mcneil1977interactiveprimer}, 
which contains the lengths, in miles, of 141 major North American rivers, 
exhibits inherent cluster structure \emph{if we allow small clusters}. 
Hopkins method and all methods that use Silverman
indicate that the data is clusterable ($p<0.05$), while all dip-based methods 
fail to reject the null hypothesis of lack of structure. Similarly, 
\emph{swiss}~\citep{mosteller1977data}, consisting of 6 
measures of socio-economic status and fertility
for 47 French-speaking nineteenth-century Swiss provinces, 
illuminated logically pre-existing structure. 
While Classic Dip considers the data as unclusterable, and Hopkins considers the data as clusterable 40\% of the time, all other tests detect clusters. 
Results support literature that economic indicators between and within countries may fall into clusters, including a richer cluster much smaller than the others ~\citep{Genicot,henderson}. 

The remaining data sets lacked previously known structure. Most tests of clusterability that weren't known or shown to be questionable in simulations provided little or no evidence of clusters. Methods based on distances or PCA concluded that \emph{cars}~\citep{ezekiel1930methods}, \emph{attitude}~\citep{chatterjee1991regression}, \emph{USArrests}~\citep{mcneil1977interactiveprimer}, \emph{trees}~~\citep{ryan1976minitab}, 
and \emph{USJudgeRatings}~\citep{Register}, 
were unclusterable. Hopkins' method agreed for 
\emph{attitude} 
and \emph{USArrests}.
Most methods (principal curve and classic) that concluded that any of these remaining data sets, without known structure, were clusterable also exhibited questionable behavior on our simulations in section~\ref{sims}.  
Classic methods may be unreliable in multiple dimensions (see section \ref{reduction}), and principal curve methods had high false positive rates in our simulations.
Classic Silverman and principal curve methods declared the \emph{USArrests}~\citep{mcneil1977interactiveprimer} and \emph{USJudgeRatings}~\citep{Register} data sets clusterable, 
%
Curiously, Hopkins considered USJudgeRatings clusterable nearly 70\% of the time, trees clusterable  18\% of the time, and cars 19\% of the time.

Overall, the methods with the most reasonable 
results include distance dip, distance Silverman, PCA dip, and PCA Silverman. Although classic Dip and Silverman methods appear to produce reasonable conclusions in some famous data sets such as \emph{iris}, they have produced counterintuitive results when classifying other real data, such as \emph{USJudgeRatings} and \emph{USArrests}. 
This finding, which supports  theory on the
unpredictability of these tests in multiple dimensions, reflects the importance and value of carefully utilizing dimensionality reduction 
in clusterability evaluation. 

\vspace{-2mm}

\section{Runtime}\label{runtime}
In this section, we discuss the runtime of the clusterability methods studied in this paper. The computational complexity of the methods is addressed in Subsection \ref{efficiency}. We report on the average observed times to complete the simulations in Subsection \ref{simulationtimes}, as well as the elapsed time to apply each method to the non-simulated data in Subsection \ref{realtimes}.

\subsection{Quantifying Efficiency} \label{efficiency}

There are significant differences in the computational complexity of clusterability techniques that render some of them impractical when the number of elements ($n$) or the dimension ($d$) is large.  
Classic Dip is linear in $n$ \citep{krause2005multimodal}. 
Hopkins, Classic Silverman, and Dip-Dist have quadratic running time in $n$. The dimensionality of the data impacts the running time of PCA-based approaches, with PCA dip having asymptotic running time of $O(nd^2+d^3)$ \citep{2015arXiv150305214E}.  Silverman-dist is bounded by a quartic function in $n$. Finally, PCA Silverman has complexity of $O(n^2 + d^2n+d^3)$.\footnote{As shown in \cite{2015arXiv150305214E}, PCA Silverman needs $O(nd^2+d^3)$ operations to calculate the principal component. Then it performs Silverman's test, which is bounded by $O(n^2)$. Therefore, the total complexity is $O(n^2 + d^2n+d^3)$.}




\subsection{Simulation times}\label{simulationtimes}
The average runtime of one execution of each clusterability method on each dataset is provided in Tables \ref{simtimes} and \ref{simtimeschaining}. Methods  utilizing the Silverman critical bandwidth test are  slower than the corresponding methods relying on the dip test, which lasted less than one hundredth of a second. As such, Dip-dist is much faster than Silv-dist, and PCA Dip is much faster than PCA Silv. The runtime of Hopkins' method generally fell in between the runtime of methods that utilize the dip test and methods that utilize the Silverman test, averaging between approximately $0.01s$ and $0.3s$.

For all simulations, the use of Silverman's test on the pairwise distances was slower than all other methods by one to three orders of magnitude, requiring between nearly $1s$ and over $8.5s$. This occurs due to the quadratic number of operations required for processing all pairwise distances, followed by the Silverman test, as discussed in Subsection \ref{efficiency}.

\begin{table*}
  \centering
  \begin{tabular}{{llrrrrrrrrrr}}
    \hline
     && \textbf{Dip} & \textbf{Silv} & \multirow{2}{*}{\textbf{Hop}}   & \textbf{Cl.} & \textbf{Cl.} & \textbf{PCA} & \textbf{PCA} & \textbf{PC}	& \textbf{PC} \\
& {\textbf{Data}}& \textbf{Dist} & \textbf{Dist} &    & \textbf{Dip} & \textbf{Silv} & \textbf{Dip} & \textbf{Silv} & \textbf{Dip}	& \textbf{Silv} \\
    \hline

\emph{1.} & 
1 cluster 2D &0.0016&0.9885&0.0130&0.0017&0.7333&0.0016&0.7221&0.0212&0.7445&\\
    \hline
\emph{2.} & 
1 cluster 3D &0.0016&0.9817&0.0128&0.0014&0.7476&0.0016&0.7227&0.0295&0.746&\\
    \hline
\emph{3.} & 
1 cluster 10D &0.0016&0.9755&0.0132&0.0014&0.8213&0.0017&0.7249&0.0742&0.7871&\\
    \hline
\emph{4.} & 
1 cluster 50D &0.0032&1.8049&0.0578&0.0018&1.8045&0.0037&0.7415&0.3119&0.728&\\
    \hline
\emph{5.} & 
1 cluster  \\& 2D with outlier &0.0016&1.0028&0.0146&0.0013&0.7429&0.0016&0.7295&0.0240&0.7444&\\
    \hline
\emph{6.} & 
1 large cluster  \\& 2D with outlier &0.0062&8.5395&0.3186&0.0014&0.8264&0.0016&0.7738&0.0834&0.8509&\\
    \hline
\emph{7.} & 
1 cluster 2D \\&with 3 outliers &0.0016&1.0331&0.0141&0.0013&0.7443&0.0016&0.7363&0.0249&0.7551&\\
    \hline
\emph{8.} & 
1 T-dist cluster \\&with df=5 &0.0021&1.8717&0.0505&0.0013&0.7908&0.0016&0.7692&0.0270&0.7937&\\
    \hline
\emph{9.} & 
1 T-dist cluster \\&with df=10 &0.0021&1.8487&0.0509&0.0013&0.7835&0.0016&0.7605&0.0270&0.7832&\\
    \hline
\emph{10.} & 
1 T-dist cluster \\&with df=15 &0.0021&1.8402&0.0508&0.0013&0.7820&0.0016&0.7564&0.0273&0.7828&\\
    \hline
\emph{11.} & 
2 separated  &0.0020&1.8538&0.0506&0.0013&0.7762&0.0016&0.7487&0.0329&0.7846&\\
&clusters 2D\\ \hline    
\emph{12.} & 
3 close   &0.0030&3.3283&0.1132&0.0013&0.7880&0.0016&0.7530&0.0267&0.7728&\\
    &clusters 2D\\ \hline
\emph{13.} & 
3 noisy  &0.0054&7.2108&0.2673&0.0014&0.8380&0.0016&0.7840&0.0479&0.8285&\\
 &clusters 2D\\   \hline
\emph{14.} & 
3 clusters 2D, \\&varied radii &0.0031&3.3522&0.1133&0.0013&0.7960&0.0016&0.7667&0.0299&0.7937&\\
    \hline
\emph{15.} & 
3 clusters 2D, \\&varied density &0.0042&5.4950&0.1899&0.0014&0.8100&0.0016&0.7769&0.0307&0.8064&\\
    \hline
\emph{16.} & 
3 separated &0.0030&3.3443&0.1139&0.0013&0.7896&0.0016&0.7587&0.0225&0.7802&\\
 &clusters 2D\\   \hline
\emph{17.} & 
3 separated  &0.0030&3.3533&0.1140&0.0014&0.8297&0.0016&0.7662&0.0324&0.7944&\\
&clusters 3D\\    \hline
\emph{18.} & 
2 separated  &0.0023&1.8576&0.0528&0.0014&0.9477&0.0017&0.7561&0.1042&0.8567&\\
&clusters 10D\\    \hline
\emph{19.} & 
4 separated &0.0048&5.5649&0.2122&0.0015&1.1694&0.0018&0.7929&0.0441&0.8227&\\
&clusters 10D\\    \hline
\emph{20.} & 
2 close &0.0086&5.5515&0.2382&0.0024&3.0199&0.0046&0.7871&0.5654&1.3229&\\
&clusters 50D\\    \hline
\emph{21.} & 
2 partially over-\\&lapping 50D &0.0087&5.5314&0.2369&0.0025&2.9757&0.0046&0.7752&0.5429&1.2955&\\
 \hline
    \emph{22.} & 
2 T-dist clust. \\&with df=5 &0.0042&5.5759&0.2007&0.0013&0.8323&0.0016&0.7985&0.0450&0.8346&\\
    \hline
\emph{23.} & 
2 T-dist clust. \\&with df=10 &0.0043&5.5694&0.2012&0.0013&0.8359&0.0016&0.7890&0.0423&0.8352&\\
    \hline
\emph{24.} & 
2 T-dist clust. \\&with df=15 &0.0042&5.5745&0.2019&0.0013&0.8381&0.0016&0.7952&0.0489&0.8363&\\
    \hline
  \end{tabular}
  \caption{ Average runtime in seconds of all methods for each simulated dataset over 1000 computations. 
  }
  \label{simtimes}
\end{table*}

\begin{table*}
  \centering
  \begin{tabular}{{llrrrrrrrrrr}}\hline
     && \textbf{Dip} & \textbf{Silv} & \multirow{2}{*}{\textbf{Hop}}   & \textbf{Cl.} & \textbf{Cl.} & \textbf{PCA} & \textbf{PCA} & \textbf{PC}	& \textbf{PC} \\
& {\textbf{Data}}& \textbf{Dist} & \textbf{Dist} &    & \textbf{Dip} & \textbf{Silv} & \textbf{Dip} & \textbf{Silv} & \textbf{Dip}	& \textbf{Silv} \\
    \hline
\emph{25.}&
Single Circle
&0.0016&0.7516&0.0026&0.0013&0.6916&0.0017&0.6822&0.0183&0.6994&\\
    \hline
\emph{26.} & 
2 concentric  &0.0022&1.8257&0.0510&0.0013&0.7524&0.0016&0.7274&0.0399&0.7653&\\
&circles\\
    \hline
\emph{27.} & 
3 concentric &0.0029&3.3041&0.1129&0.0013&0.7760&0.0016&0.7431&0.0503&0.7869&\\
&circles\\
    \hline
\emph{28.} & 
5 concentric  &0.0062&8.4360&0.3162&0.0014&0.8157&0.0017&0.7628&0.0736&0.8357&\\
&circles\\
    \hline
\emph{29.} &
Single line
&0.0025&2.0360&0.0595&0.0017&0.8681&0.0019&0.8393&0.1624&0.1608&\\ 
    \hline
\emph{30.} & 
2 parallel lines &0.0044&5.5498&0.2024&0.0013&0.8232&0.0016&0.7801&0.0687&0.8394&\\  \hline
\emph{31.}&
Line and 
&0.0022&1.8136&0.0524&0.0016&0.7590&0.0018&0.7417&0.0544&0.7971&\\ 
&circle\\

    \hline
  \end{tabular}
  \caption{ Average run time in seconds of all methods for each simulated dataset with chaining structure over 1000 computations. 
  }
  \label{simtimeschaining}
\end{table*}

\subsection{Runtime for non-simulated data}\label{realtimes}
The relative runtime for non-simulated data was comparable to that of the simulations, and confirmed expectations based on the methods' computational complexity. 
Hopkins is slower than dist-dip and PCA dip but faster than Silverman methods. PCA dip is similar to distance dip, both of which ran in one to three thousandths of a second. Reducing the data using PCA and then testing with the Silverman critical bandwidth test took about seven-tenths to eight-tenths of a second on non-simulated data. By contrast, running Silverman's test on the set of pairwise distances takes much longer, nearly ten seconds for the largest dataset, {\it faithful}, compared to about three seconds for the two datasets that are about half of the size. 

\begin{table*}
  \centering
  \begin{tabular}{{lrrrrrrrrrrrr}}
    \hline
     &&&\textbf{Dip} & \textbf{Silv} & \multirow{2}{*}{\textbf{Hop}}   & \textbf{Cl.} & \textbf{Cl.} & \textbf{PCA} & \textbf{PCA} & \textbf{PC}	& \textbf{PC} \\
 {\textbf{Data}}& n&d& \textbf{Dist} & \textbf{Dist} &    & \textbf{Dip} & \textbf{Silv} & \textbf{Dip} & \textbf{Silv} & \textbf{Dip}	& \textbf{Silv} \\
    \hline
Faithful&272&2&0.0113&9.8536&0.3696&0.0020&0.8126&0.0015&0.7910&0.4809&1.2679&\\
    \hline
Iris&150&4&0.0030&3.2233&0.1136&0.0015&0.8166&0.0020&0.7305&0.0270&0.7530&\\
    \hline
Rivers&141&1&0.0030&3.0371&0.0970&0.0010&0.7530&0.0015&0.8021&0.1386&0.9036&\\
    \hline
Swiss&47&6&0.0020&0.9872&0.0099&0.0015&0.8286&0.0015&0.7285&0.0605&0.8351&\\
    \hline
Attitude&30&7&0.0015&0.8063&0.0048&0.0015&0.7770&0.0015&0.7410&0.0555&0.7475&\\
    \hline
Cars&50&2&0.0015&0.9537&0.0126&0.0010&0.7270&0.0020&0.7856&0.0070&0.7435&\\
    \hline
Trees&31&3&0.0010&0.7906&0.0049&0.0010&0.7780&0.0015&0.7350&0.0110&0.6890&\\
    \hline
Ratings&43&12&0.0015&0.9156&0.0093&0.0010&0.8071&0.0015&0.7090&0.0405&0.7846&\\
    \hline
Arrests&50&4&0.0015&1.0182&0.0129&0.0015&0.7921&0.0015&0.7595&0.0085&0.7565&\\
    \hline
  \end{tabular}
  \caption{ Runtime in seconds of each method on each non-simulated dataset. Recorded time for Hopkins is the average of 100 runs. The numbers of observations and number of features of the data are denoted by {\it n} and {\it d}.
  }
\end{table*}

\section{Discussion}\label{discuss}

Though many approaches to clusterability evaluation have been previously proposed, they vary radically and often result in different conclusions. Here, we perform an extensive analysis of a variety of clusterability methods, identifying which are most effective as well as when certain measures are better suited than others based on the needs of the application at hand. 
%
%
While other notions of clusterability may also warrant investigation, our paper is the most comprehensive study to date. We compare several approaches, which apply either spatial randomness tests to the original data or multimodality tests to one-dimensional reductions of the data. Extensive simulations allow us to identify effective approaches, as well as differentiate amongst them. Notably, experiments on real data sets parallel the conclusions of our simulations.  
%
%
Our findings indicate that spatial randomness tests and 
multimodality tests on 
one-dimensional reductions 
are frequently effective 
at classifying data sets by their level of clusterability. Both clusterable and unclusterable data sets were identified as such in most simulations. 

Methods perform differently according to dimension, treatment of outliers, and shape and separability of clusters. 
Distance-based methods perform well in most scenarios. PCA methods detect structure  in data with two or three clusters and does not detect spurious clusters in data with a single cluster. 
However, in low dimensions, PCA power is lower than for distance-based methods, {
{and PCA performs poorly for non-linear data}. 
Outliers are treated as clusters by all variations of Silverman and the Hopkins statistic. The Hopkins statistic loses signal when clusters touch or overlap.  
Classical methods are inappropriate in multiple dimensions and for chaining data. 
Finally, principal curve methods were highly problematic on both simulated and non-simulated data sets. Even on the famous, well-defined data set {\it faithful}, the principal curve failed to converge after 1000 iterations. Principal curve methods also had excessive false positives on simulated data generated from a single distribution.
We summarize several qualitative criteria that can be used to select a suitable clusterability measure for a given application. 
{
Quantitative comparison based on the efficiency of these methods could also be integrated, particularly when data is large in number of observations or dimensions or both. See Section~\ref{runtime} for a comparison of the methods considered in this analysis based on their computational complexity and empirical runtime on simulated and non-simulated data. 


}}

\begin{itemize}[leftmargin=*]

\item \textbf{False Positives:}  Methods proclaiming to discover cluster structure when none is present have excessive false positives (Type I error) and are considered statistically invalid. Reducing data using the principle curve 
consistently yielded inflated Type I error rates and is not recommended.



\item \textbf{Outliers/Small Clusters:}  
Clusterability measures vary 
in their treatment of sparse distant points. 
Hopkins and Silverman-based methods treat the points as small clusters; Dip-based methods exhibit outlier robustness. 




\item \textbf{Chaining Data:} Dip-dist was the only method that consistently performed well on chaining-type data, 
identifying both clusterable and unclusterable structures of these types.

\item \textbf{High Dimensionality:} We tested data sets on up to 50 dimensions. In our experiments, PCA dip, PCA Silverman, and Dip-dist 
did well and were reasonably efficient, suggesting that these methods may be better suited to high dimensional data than the other techniques considered in this analysis. 
\end{itemize}






\vspace{2mm}

While our results suggest that some of the methods considered here work well for data of reasonably high dimension, for very high dimensional data (particularly when the number of dimensions is much greater than the number of elements), additional investigation is desirable. It is possible that simply modifying the data reduction method, such as by using Sparse PCA \citep{zou2006sparsePCA}, may be sufficient. These avenues of investigation are left for future work. 
{
{However, we would expect that Dist-dip would be the most efficient method, followed by Hopkins, for such data based on the theoretical complexity discussed in Subsection \ref{efficiency}.}}

{
{The choice of dimension reduction method should be made carefully. 
For example, if applicable, an appropriate}} distance metric is crucial for proper analysis in both cluster analysis and clusterability evaluation. For the purpose of this paper, we focus on Euclidian distance, which is the most common metric. Investigation of other metrics, {
{while a popular topic of recent interest \cite{LU20181},}} is left for future research. {
{Similarly, if the data is known to be highly non-linear, then the user may not wish to use PCA. Other data reduction methods for non-linear data will be explored in future work.}}




\section{Conclusions}

The application of clustering algorithms presupposes the existence of cluster structure. Clustering techniques tend to produce some partition for any given data set, which can lead to invalid conclusions when the data is  unclusterable. Consequently, we advocate for the integration of clusterability into cluster analysis, allowing users to determine whether clustering is appropriate for the given data before proceeding with further processing. 
%


A succinct, practical summary of our results is shown in the table below. The table includes only methods that were found to be effective in our simulations, and allows users to select from amongst these methods on the basis of three simple metrics: whether small clusters should be treated as outlier (do we want an outlier robust measure, or one that allows for small clusters?), whether the target clustering may consists of clusters with chaining structure, 
or is high dimensional (up to $50$ dimensions).
The following table focuses on qualitative analysis,  see Section~\ref{runtime} for runtime analysis. 

\begin{center}
\begin{tabular}{ l  c   c  c  c}
& Outliers/Small Clusters &   Chaining Data &   High Dimensional\\
\hline
PCA dip &  small clusters & x & $\checkmark$\\ 
\hline

PCA Silverman &  small clusters & x & $\checkmark$\\  
\hline

Dist dip &  outlier robust & $\checkmark$ & $\checkmark$\\
\hline

Dist Silverman &  outlier robust & x &x\\
\hline

Hopkins &  small clusters & x & x\\

\hline
\end{tabular}
\end{center}


Please see Section~\ref{discuss} for a thorough discussion. 

We look forward to the widespread application of clusterability tests as part of the clustering process. 
{
{As of the writing of the present manuscript, implementation of the clusterability methods described in this paper is in progress. To this end, two macros were written to conduct multimodality tests in {\it SAS}, as the procedures were previously unavailable in {\it SAS} \cite{nevillebrownstein}. Software development is in progress to facilitate user-friendly applications of the clusterability methods in  common statistical software {\it SAS} and {\it R}.  Documentation of these clusterability algorithms is also in development and will be published in future manuscripts.   
} }

We close with the following quote to remind of the importance of testing for clusterability before proceeding with further -- potentially unnecessary -- cluster analysis tasks.

\vspace{3mm}

\begin{quote}
``There is nothing so useless as doing efficiently that which should not be done at all.'' -Peter F. Drucker
\end{quote}

\section{Acknowledgments}
This research did not receive any specific grant from funding agencies in the public, commercial, or not-for-profit sectors.

\appendix

\bibliographystyle{elsarticle-num} 
\bibliography{clustering}

\begin{thebibliography}{10}
\expandafter\ifx\csname url\endcsname\relax
  \def\url#1{\texttt{#1}}\fi
\expandafter\ifx\csname urlprefix\endcsname\relax\def\urlprefix{URL }\fi
\expandafter\ifx\csname href\endcsname\relax
  \def\href#1#2{#2} \def\path#1{#1}\fi

\bibitem{Kleinberg}
J.~Kleinberg, An impossibility theorem for clustering, in: Advances in Neural
  Information Processing Systems, 2003, pp. 463--470.

\bibitem{NIPS2008}
M.~Ackerman, S.~Ben-David, Measures of clustering quality: A working set of
  axioms for clustering, in: Advances in Neural Information Processing Systems,
  2008, pp. 121--128.

\bibitem{NIPS2010}
M.~Ackerman, S.~Ben-David, D.~Loker, Towards property-based classification of
  clustering paradigms, in: Advances in Neural Information Processing Systems,
  2010, pp. 10--18.

\bibitem{COLT2010}
M.~Ackerman, S.~{Ben-David}, D.~Loker, Characterization of linkage-based
  clustering., in: Conference on Learning Theory (COLT), 2010, pp. 270--281.

\bibitem{IJCAI2011}
M.~Ackerman, S.~Ben-David, Discerning linkage-based Algorithms among
  hierarchical clustering methods, in: IJCAI Proceedings-International Joint
  Conference on Artificial Intelligence, Vol.~22, 2011, p. 1140.

\bibitem{weighted}
M.~Ackerman, S.~Ben-David, S.~Branzei, D.~Loker, Weighted clustering., in:
  Proceedings of the AAAI Conference (Association for the Advancement of
  Artificial Intelligence), 2012, pp. 858--863.

\bibitem{hennig2015true}
C.~Hennig, What are the true clusters?, Pattern Recognition Letters 64 (2015)
  53--62.

\bibitem{xu2005survey}
R.~Xu, D.~Wunsch, Survey of clustering algorithms, Neural Networks, IEEE
  Transactions on 16~(3) (2005) 645--678.

\bibitem{ackerman2014incremental}
M.~Ackerman, S.~Dasgupta, Incremental clustering: the case for extra clusters,
  in: Advances in Neural Information Processing Systems, 2014, pp. 307--315.

\bibitem{Milligan}
G.~Milligan, A {M}onte {C}arlo study of thirty internal criterion measures for
  cluster analysis, Psychometrika 46~(2) (1981) 187--199.

\bibitem{ackerman2009clusterability}
M.~Ackerman, S.~Ben-David, Clusterability: A theoretical study, Proceedings of
  AISTATS-09, JMLR: W\&CP 5~(1-8) (2009) 53.

\bibitem{ben2015computational}
S.~Ben-David, Computational feasibility of clustering under clusterability
  assumptions, arXiv preprint arXiv:1501.00437.

\bibitem{oligarchies}
M.~Ackerman, S.~{Ben-David}, D.~Loker, S.~Sabato, Clustering oligarchies,
  Proceedings of AISTATS-09, Journal of Machine Learning Research: W\&CP
  31~(66–74).

\bibitem{wold}
S.~Wold, K.~Esbensen, P.~Geladi,
  \href{http://www.sciencedirect.com/science/article/pii/0169743987800849}{Principal
  component analysis}, Chemometrics and intelligent laboratory systems 2~(1)
  (1987) 37 -- 52.
\newline\urlprefix\url{http://www.sciencedirect.com/science/article/pii/0169743987800849}

\bibitem{jolliffe2002pca}
I.~Jolliffe, \href{https://books.google.com/books?id=\_olByCrhjwIC}{Principal
  Component Analysis}, Springer Series in Statistics, Springer, 2002.
\newline\urlprefix\url{https://books.google.com/books?id=\_olByCrhjwIC}

\bibitem{zha2001NIPS}
H.~Zha, X.~He, C.~Ding, M.~Gu, H.~D. Simon,
  \href{http://papers.nips.cc/paper/1992-spectral-relaxation-for-k-means-clustering.pdf}{Spectral
  relaxation for $k$-means clustering}, in: T.~G. Dietterich, S.~Becker,
  Z.~Ghahramani (Eds.), Advances in Neural Information Processing Systems 14,
  MIT Press, 2002, pp. 1057--1064.
\newline\urlprefix\url{http://papers.nips.cc/paper/1992-spectral-relaxation-for-k-means-clustering.pdf}

\bibitem{Ding2004}
C.~Ding, X.~He, \href{http://doi.acm.org/10.1145/1015330.1015408}{K-means
  clustering via principal component analysis}, in: Proceedings of the
  Twenty-first International Conference on Machine Learning, ICML '04, ACM, New
  York, NY, USA, 2004, pp. 29--.
\newline\urlprefix\url{http://doi.acm.org/10.1145/1015330.1015408}

\bibitem{varmuza}
K.~Varmuza, P.~Filzmoser, Introduction to multivariate statistical analysis in
  chemometrics, CRC Press, 2009.

\bibitem{Huber1985}
P.~J. Huber, \href{http://www.jstor.org/stable/2241175}{Projection pursuit},
  The Annals of Statistics 13~(2) (1985) 435--475.
\newline\urlprefix\url{http://www.jstor.org/stable/2241175}

\bibitem{krause2005multimodal}
A.~Krause, V.~Liebscher, Multimodal projection pursuit using the dip statistic,
  Preprint-Reihe Mathematik 13.

\bibitem{HastieStuetzle}
T.~Hastie, W.~Stuetzle,
  \href{http://amstat.tandfonline.com/doi/abs/10.1080/01621459.1989.10478797}{Principal
  curves}, Journal of the American Statistical Association 84~(406) (1989)
  502--516.
\newline\urlprefix\url{http://amstat.tandfonline.com/doi/abs/10.1080/01621459.1989.10478797}

\bibitem{Tibshirani1992}
R.~Tibshirani, \href{http://dx.doi.org/10.1007/BF01889678}{Principal curves
  revisited}, Statistics and Computing 2~(4) (1992) 183--190.
\newline\urlprefix\url{http://dx.doi.org/10.1007/BF01889678}

\bibitem{goslee}
S.~C. Goslee, \href{http://www.jstor.org/stable/40540359}{Correlation analysis
  of dissimilarity matrices}, Plant Ecology 206~(2) (2010) 279--286.
\newline\urlprefix\url{http://www.jstor.org/stable/40540359}

\bibitem{XuEdwardHansonRestrepo}
X.~Ling, E.~J. Bedrick, T.~Hanson, C.~Restrepo,
  \href{https://login.proxy.lib.fsu.edu/login?url=http://search.ebscohost.com/login.aspx?direct=true&db=a9h&AN=93452761&site=eds-live}{A
  comparison of statistical tools for identifying modality in body mass
  distributions.}, Journal of Data Science 12 (2014) 175 -- 196.
\newline\urlprefix\url{https://login.proxy.lib.fsu.edu/login?url=http://search.ebscohost.com/login.aspx?direct=true&db=a9h&AN=93452761&site=eds-live}

\bibitem{dip_means}
A.~Kalogeratos, A.~Likas, Dip-means: an incremental clustering method for
  estimating the number of clusters, in: F.~Pereira, C.~J.~C. Burges,
  L.~Bottou, K.~Q. Weinberger (Eds.), Advances in Neural Information Processing
  Systems 25, Curran Associates, Inc., 2012, pp. 2393--2401.

\bibitem{LuHayesNobelMarron}
Y.~Liu, D.~N. Hayes, A.~Nobel, J.~S. Marron,
  \href{http://dx.doi.org/10.1198/016214508000000454}{Statistical significance
  of clustering for high-dimension, low–sample size data}, Journal of the
  American Statistical Association 103~(483) (2008) 1281--1293.
\newline\urlprefix\url{http://dx.doi.org/10.1198/016214508000000454}

\bibitem{sigtest}
M.~Shahbaba, S.~Beheshti, Efficient unimodality test in clustering by signature
  testing, in: 2014 IEEE International Conference on Acoustics, Speech and
  Signal Processing (ICASSP), 2014, pp. 8282--8286.
\newblock \href {http://dx.doi.org/10.1109/ICASSP.2014.6855216}
  {\path{doi:10.1109/ICASSP.2014.6855216}}.

\bibitem{Helgeson}
E.~S. {Helgeson}, E.~{Bair}, {Non-parametric cluster significance testing with
  reference to a unimodal null distribution}, ArXiv e-prints\href
  {http://arxiv.org/abs/1610.01424} {\path{arXiv:1610.01424}}.

\bibitem{kimes}
P.~K. Kimes, Y.~Liu, D.~Neil~Hayes, J.~S. Marron,
  \href{http://dx.doi.org/10.1111/biom.12647}{Statistical significance for
  hierarchical clustering}, Biometrics (2017) n/a--n/a.
\newline\urlprefix\url{http://dx.doi.org/10.1111/biom.12647}

\bibitem{hartiganDip}
J.~A. Hartigan, P.~M. Hartigan,
  \href{http://dx.doi.org/10.1214/aos/1176346577}{The Dip Test of Unimodality},
  Ann. Statist. 13~(1) (1985) 70--84.
\newline\urlprefix\url{http://dx.doi.org/10.1214/aos/1176346577}

\bibitem{dipprogramjrssc}
P.~M. Hartigan, \href{http://www.jstor.org/stable/2347485}{Algorithm AS 217:
  Computation of the Dip Statistic to Test for Unimodality}, Journal of the
  Royal Statistical Society. Series C (Applied Statistics) 34~(3) (1985)
  320--325.
\newline\urlprefix\url{http://www.jstor.org/stable/2347485}

\bibitem{citediptestR}
M.~Maechler, R, l package version 0.75-7 (2015).
\newline\urlprefix\url{http://CRAN.R-project.org/package=diptest}

\bibitem{silverman}
B.~W. Silverman, \href{http://www.jstor.org/stable/2985156}{Using kernel
  density estimates to investigate multimodality}, Journal of the Royal
  Statistical Society. Series B (Methodological) 43~(1) (1981) pp. 97--99.
\newline\urlprefix\url{http://www.jstor.org/stable/2985156}

\bibitem{NAP1360}
N.~R. Council,
  \href{http://www.nap.edu/catalog/1360/discriminant-analysis-and-clustering}{Discriminant
  analysis and clustering}, The National Academies Press, Washington, DC, 1988.
\newline\urlprefix\url{http://www.nap.edu/catalog/1360/discriminant-analysis-and-clustering}

\bibitem{HartiganTheory}
J.~Hartigan, \href{http://dx.doi.org/10.1007/BF01908064}{Statistical theory in
  clustering}, Journal of Classification 2~(1) (1985) 63--76.
\newline\urlprefix\url{http://dx.doi.org/10.1007/BF01908064}

\bibitem{Wells1978}
D.~R. Wells, \href{http://www.jstor.org/stable/2958871}{A monotone unimodal
  distribution which is not central convex unimodal}, The Annals of Statistics
  6~(4) (1978) 926--931.
\newline\urlprefix\url{http://www.jstor.org/stable/2958871}

\bibitem{citeR}
{R Core Team}, \href{https://www.R-project.org}{R: A language and environment
  for statistical computing}, R Foundation for Statistical Computing, Vienna,
  Austria (2015).
\newline\urlprefix\url{https://www.R-project.org}

\bibitem{Ahmed2012}
M.~O. Ahmed, G.~Walther,
  \href{http://www.sciencedirect.com/science/article/pii/S0167947312001028}{Investigating
  the multimodality of multivariate data with principal curves}, Computational
  Statistics and Data Analysis 56~(12) (2012) 4462 -- 4469.
\newline\urlprefix\url{http://www.sciencedirect.com/science/article/pii/S0167947312001028}

\bibitem{citeSilvR}
F.~Schwaiger, H.~Holzmann,
  \href{https://www.uni-marburg.de/fb12/stoch/research/rpackage?language_sync=1}{Package
  ‘silvermantest’.} (2013).
\newline\urlprefix\url{https://www.uni-marburg.de/fb12/stoch/research/rpackage?language_sync=1}

\bibitem{HallYork}
P.~Hall, M.~York,
  \href{http://www3.stat.sinica.edu.tw/statistica/oldpdf/A11n28.pdf}{{On the
  calibration of Silverman's test for multimodality}}, Statistica Sinica 11
  (2001) 515--536.
\newline\urlprefix\url{http://www3.stat.sinica.edu.tw/statistica/oldpdf/A11n28.pdf}

\bibitem{ChengHall}
M.-Y. Cheng, P.~Hall,
  \href{http://dx.doi.org/10.1111/1467-9868.00141}{Calibrating the excess mass
  and dip tests of modality}, Journal of the Royal Statistical Society: Series
  B (Statistical Methodology) 60~(3) (1998) 579--589.
\newline\urlprefix\url{http://dx.doi.org/10.1111/1467-9868.00141}

\bibitem{citeprincurvR}
S.~original by Trevor Hastie R port~by Andreas
  Weingessel~<Andreas.Weingessel@ci.tuwien.ac.at>, R, l package version 1.1-12
  (2013).
\newline\urlprefix\url{http://CRAN.R-project.org/package=princurve}

\bibitem{hopkins_original}
B.~Hopkins, J.~G. Skellam, \href{http://www.jstor.org/stable/42907238}{A new
  method for determining the type of distribution of plant individuals}, Annals
  of Botany 18~(70) (1954) 213--227.
\newline\urlprefix\url{http://www.jstor.org/stable/42907238}

\bibitem{LawsonJurs1990}
R.~G. Lawson, P.~C. Jurs, \href{http://dx.doi.org/10.1021/ci00065a010}{New
  index for clustering tendency and its application to chemical problems},
  Journal of Chemical Information and Computer Sciences 30~(1) (1990) 36--41.
\newline\urlprefix\url{http://dx.doi.org/10.1021/ci00065a010}

\bibitem{Dubes1987}
R.~C. Dubes, G.~Zeng, \href{http://dx.doi.org/10.1007/BF01890074}{A test for
  spatial homogeneity in cluster analysis}, Journal of Classification 4~(1)
  (1987) 33--56.
\newline\urlprefix\url{http://dx.doi.org/10.1007/BF01890074}

\bibitem{banerjeeDave}
A.~Banerjee, R.~N. Dave, Validating clusters using the Hopkins statistic, in:
  2004 IEEE International Conference on Fuzzy Systems (IEEE Cat. No.04CH37542),
  Vol.~1, 2004, pp. 149--153 vol.1.
\newblock \href {http://dx.doi.org/10.1109/FUZZY.2004.1375706}
  {\path{doi:10.1109/FUZZY.2004.1375706}}.

\bibitem{citeHopkinsR}
L.~YiLan, Z.~RuTong, R, l package version 1.4 (2015).
\newline\urlprefix\url{http://CRAN.R-project.org/package=clustertend}

\bibitem{Epter}
S.~Epter, M.~Krishnamoorthy, M.~Zaki, Clusterability detection and initial seed
  selection in large datasets, in: The International Conference on Knowledge
  Discovery in Databases, Vol.~7, 1999.

\bibitem{awasthi2010stability}
P.~Awasthi, A.~Blum, O.~Sheffet, Stability yields a PTAS for k-median and
  k-means clustering, in: Foundations of Computer Science (FOCS), 2010 51st
  Annual IEEE Symposium on, IEEE, 2010, pp. 309--318.

\bibitem{awasthi2012center}
P.~Awasthi, A.~Blum, O.~Sheffet, Center-based clustering under perturbation
  stability, Information Processing Letters 112~(1) (2012) 49--54.

\bibitem{balcan2009approximate}
M.-F. Balcan, A.~Blum, A.~Gupta, Approximate clustering without the
  approximation, in: Proceedings of the twentieth Annual ACM-SIAM Symposium on
  Discrete Algorithms, Society for Industrial and Applied Mathematics, 2009,
  pp. 1068--1077.

\bibitem{Swamy}
R.~Ostrovsky, Y.~Rabani, L.~Schulman, C.~Swamy, The effectiveness of
  {L}loyd-type methods for the $k$-means problem, in: Foundations of Computer
  Science, 2006. FOCS'06. 47th Annual IEEE Symposium on, 2006, pp. 165--176.

\bibitem{Szczubialka}
K.~Szczubialka, J.~Verdú-Andrés, D.~Massart,
  \href{http://www.sciencedirect.com/science/article/pii/S0169743997000774}{A
  new method of detecting clustering in the data}, Chemometrics and Intelligent
  Laboratory Systems 41~(2) (1998) 145 -- 160.
\newline\urlprefix\url{http://www.sciencedirect.com/science/article/pii/S0169743997000774}

\bibitem{NordhaugMyhre2018491}
J.~N. Myhre, K.~Øyvind Mikalsen, S.~Løkse, R.~Jenssen,
  \href{https://www.sciencedirect.com/science/article/pii/S0031320317304776}{Robust
  clustering using a kNN mode seeking ensemble}, Pattern Recognition 76 (2018)
  491 -- 505.
\newline\urlprefix\url{https://www.sciencedirect.com/science/article/pii/S0031320317304776}

\bibitem{celeux1995gaussian}
G.~Celeux, G.~Govaert, Gaussian parsimonious clustering models, Pattern
  recognition 28~(5) (1995) 781--793.

\bibitem{dasgupta1999learning}
S.~Dasgupta, Learning mixtures of Gaussians, in: Foundations of computer
  science, 1999. 40th annual symposium on, IEEE, 1999, pp. 634--644.

\bibitem{fisher1936use}
R.~A. Fisher, The use of multiple measurements in taxonomic problems, Annals of
  eugenics 7~(2) (1936) 179--188.

\bibitem{hardle1991smoothing}
W.~H{\"a}rdle, Smoothing techniques: with implementation in S, Springer Science
  \& Business Media, 1991.

\bibitem{azzalini1990look}
A.~Azzalini, A.~W. Bowman, A look at some data on the Old Faithful geyser,
  Applied Statistics (1990) 357--365.

\bibitem{AlvarezPena}
D.~Pena, A.~Alvarez,
  \href{https://ideas.repec.org/p/cte/wsrepe/ws130706.html}{Recombining
  partitions via unimodality tests}, DES - Working Papers. Statistics and
  Econometrics. WS ws130706, Universidad Carlos III de Madrid. Departamento de
  Estadística (Mar 2013).
\newline\urlprefix\url{https://ideas.repec.org/p/cte/wsrepe/ws130706.html}

\bibitem{mcneil1977interactiveprimer}
D.~R. McNeil, Interactive data analysis: a practical primer, John Wiley \&
  Sons, 1977.

\bibitem{mosteller1977data}
F.~Mosteller, J.~W. Tukey, Data analysis and regression: a second course in
  statistics., Addison-Wesley Series in Behavioral Science: Quantitative
  Methods.

\bibitem{ezekiel1930methods}
M.~Ezekiel, methods of correlation analysis. 427 pp., illus, New York and
  London.

\bibitem{chatterjee1991regression}
S.~Chatterjee, B.~Price, Regression analysis by example, John Wiley \& Sons,
  1991.

\bibitem{ryan1976minitab}
T.~A. Ryan, B.~L. Joiner, B.~F. Ryan, et~al., Minitab student handbook, Duxbury
  Press, 1976.

\bibitem{Register}
{John Hartigan}, {New Haven Register} (1977).

\bibitem{Genicot}
G.~Genicot, D.~Ray, \href{http://www.nber.org/papers/w19976}{Aspirations and
  inequality}, Working Paper 19976, National Bureau of Economic Research (March
  2014).
\newline\urlprefix\url{http://www.nber.org/papers/w19976}

\bibitem{henderson}
D.~J. Henderson, C.~F. Parmeter, R.~R. Russell,
  \href{http://dx.doi.org/10.1002/jae.1023}{Modes, weighted modes, and
  calibrated modes: evidence of clustering using modality tests}, Journal of
  Applied Econometrics 23~(5) (2008) 607--638.
\newline\urlprefix\url{http://dx.doi.org/10.1002/jae.1023}

\bibitem{2015arXiv150305214E}
T.~{Elgamal}, M.~{Hefeeda}, {Analysis of PCA algorithms in distributed
  environments}, ArXiv e-prints\href {http://arxiv.org/abs/1503.05214}
  {\path{arXiv:1503.05214}}.

\bibitem{zou2006sparsePCA}
H.~Zou, T.~Hastie, R.~Tibshirani,
  \href{http://dx.doi.org/10.1198/106186006X113430}{Sparse principal component
  analysis}, Journal of Computational and Graphical Statistics 15~(2) (2006)
  265--286.
\newline\urlprefix\url{http://dx.doi.org/10.1198/106186006X113430}

\bibitem{LU20181}
J.~Lu, R.~Wang, A.~Mian, A.~Kumar, S.~Sarkar,
  \href{http://www.sciencedirect.com/science/article/pii/S0031320317304399}{Distance
  metric learning for pattern recognition}, Pattern RecognitionD, listance
  Metric Learning for Pattern Recognition.
\newline\urlprefix\url{http://www.sciencedirect.com/science/article/pii/S0031320317304399}

\bibitem{nevillebrownstein}
Z.~{Neville}, N.~{Brownstein}, {Macros to conduct tests of multimodality in
  SAS}, ArXiv e-prints\href {http://arxiv.org/abs/1802.00151}
  {\path{arXiv:1802.00151}}.

\end{thebibliography}





\end{document}